%% file: main.tex
\documentclass[runningheads]{llncs}
\usepackage{graphicx}
\usepackage{booktabs}
\usepackage{hyperref}
\usepackage{multirow}
\usepackage{subfig}
\usepackage{amsmath}
\usepackage{algorithm}
\usepackage{algpseudocode}
\usepackage[capitalize]{cleveref}
\usepackage{tikz}
\def\checkmark{\tikz\fill[scale=0.4](0,.35) -- (.25,0) -- (1,.7) -- (.25,.15) -- cycle;}

\DeclareSymbolFont{toneletters}{T1}{\familydefault}{m}{it}
\SetSymbolFont{toneletters}{bold}{T1}{\familydefault}{bx}{it}

\begin{document}
\title{GraphKD: Exploring Knowledge Distillation Towards Document Object Detection with Structured Graph Creation}
\titlerunning{GraphKD}
%
\author{Ayan Banerjee\inst{1}\orcidID{0000-0002-0269-2202} \and
Sanket Biswas\inst{1}\orcidID{0000-0001-6648-8270} \and
Josep Lladós\inst{1}\orcidID{0000-0002-4533-4739} \and
Umapada Pal\inst{2}\orcidID{0000-0002-5426-2618}}
\authorrunning{A.Banerjee et al.}
\institute{Computer Vision Center \& Computer Science Department  \\
              Universitat Autònoma de Barcelona, Spain \\
              \email{\{abanerjee, sbiswas, josep\}@cvc.uab.es} 
              \and
              CVPR Unit, Indian Statistical Institute, Kolkata, India \\ 
              \email{umapada@isical.ac.in}}
\maketitle              
\input{./tex/abs.tex}
\section{Introduction}
\label{s:intro}
\input{./tex/intro.tex}

\section{Related Work}
\label{s:sota}
\input{./tex/sota.tex}

\section{Method}
\label{s:method}
\input{./tex/method.tex}

\section{Experimental Evaluation}
\label{s:experiments}
\input{./tex/experiments.tex}

\section{Conclusion}
\label{s:conclusion}
\input{./tex/conclusion.tex}

\section*{Acknowledgment}
This work has been partially supported by the Spanish project PID2021-126808OB-I00, the Catalan project 2021 SGR 01559 and the PhD Scholarship from AGAUR (2021FIB-10010). The Computer Vision Center is part of the CERCA Program/Generalitat de Catalunya.

\bibliographystyle{splncs04}
\bibliography{main}

\newpage
\input{tex/supply}
\end{document}

%% file: tex/abs.tex
\begin{abstract}
Object detection in documents is a key step to automate the structural elements identification process in a digital or scanned document through understanding the hierarchical structure and relationships between different elements. Large and complex models, while achieving high accuracy, can be computationally expensive and memory-intensive, making them impractical for deployment on resource constrained devices. Knowledge distillation allows us to create small and more efficient models that retain much of the performance of their larger counterparts. Here we present a graph-based knowledge distillation framework to correctly identify and localize the document objects in a document image. Here, we design a structured graph with nodes containing proposal-level features and edges representing the relationship between the different proposal regions. Also, to reduce text bias an adaptive node sampling strategy is designed to prune the weight distribution and put more weightage on non-text nodes. We encode the complete graph as a knowledge representation and transfer it from the teacher to the student through the proposed distillation loss by effectively capturing both local and global information concurrently. Extensive experimentation on competitive benchmarks demonstrates that the proposed framework outperforms the current state-of-the-art approaches. The code will be available at: \href{https://github.com/ayanban011/GraphKD}{github.com/ayanban011/GraphKD}

\keywords{Knowledge Distillation  \and Document Object Detection \and Graph Neural Network.}
\end{abstract}

%% file: tex/intro.tex
Document Object Detection (DOD) has indeed become an essential task in Document Understanding (DU) because any task related to document understanding entails the need to obtain a structured representation that helps to distinguish between text, images, tables, headings, footers, and other design elements \cite{mathur2023layerdoc,pfitzmann2022doclaynet}. The main goal of DOD is to understand the structure and content of a document, typically as a precursor to more detailed processing or analysis. For instance, in the field of Optical Character Recognition (OCR) \cite{fateh2023enhancing}, DOD can be used to determine which parts of a page contain text that needs to be recognized, versus which parts might contain pictures or other non-textual elements. Similarly, in Document Retrieval \cite{coquenet2023dan}, Key Information Extraction \cite{stanislawek2021kleister}, and Visual Question Answering \cite{wu2022region}, DOD helps to localize the region where the key information or the answer has been located. For this reason, throughout the last decade, remarkable progress has been observed, starting from the convolution based algorithms (e.g. Faster-RCNN \cite{sun2019faster}, Mask R-CNN \cite{zhong2023hybrid}, RetinaNet \cite{lin2021keyword}, and so on) to the large-scale multi-modal transformers \cite{huang2022layoutlmv3,tang2023unifying}. With the increasing complexity of document layouts, the model complexity and the number of model parameters have also increased. Although the conventional approaches \cite{banerjee2023swindocsegmenter,li2022dit} demonstrate significant performance, we cannot use them in our edge devices (we are penalized by the computation cost). On the other hand, tiny networks can be used for edge devices but we are penalized by the object detection performance (i.e. efficiency). In order to solve this trade-off between memory and efficiency we proposed a graph-based knowledge distillation technique where we trained large networks in the GPUs and encoded their learned features in a graph data structure to transfer it to the tiny networks by optimizing a distillation loss, so that we can use the same in the edge devices. According to our best knowledge, this is the first attempt to perform knowledge distillation (KD) in order to solve the Document Object Detection (DOD) task.

However, KD in object detection is not easy because we need to consider the spatial location of the multiple objects along with their scale variation and imbalanced distributions. Conventional KD techniques \cite{gou2023multi,gong2023adaptive,lin2023supervised} are often fail to deal with the feature imbalance problem and also unable to identify the missing instance-level relations as they perform the distillation through one of the following techniques. 1) \textit{Logit-based:} here we distill only the logits to the final softmax layer. This technique always loses the fine-grained information as it doesn't consider the feature maps of the intermediate layers. 2) \textit{Feature-based:} It performs layer-to-layer feature distillation but suffers with feature alignment problems. So, only homogeneous distillation (e.g. ResNet101-ResNet50) is possible. 3) \textit{Hybrid:} It performs a layer-to-layer distillation of the feature maps as well as logit-to-logit distillation and the final softmax layer but it diminishes the transferability. In order to tackle these issues, we construct a structured instance graph where we collect the regional features from the RoI instances of the RPN and store them in the graph nodes. On the other hand, the edges represent the relations between nodes and are measured by their feature similarity. So, the nodes help to overcome the feature imbalance problem and edges excavate the missing instance relation. Not only that graphs also preserved the structural information during the embedding transfer and adapted easily to the topological structure between teacher and student even with different widths and depths enabling us to perform heterogeneous distillation (EfficientNet to ResNet, MobileNet to ResNet, and so on).

Even the distillation through the graph for the documents creates some additional issues. Initially, a notable challenge arises due to the large number of text nodes providing text bias which is propagated as a noise for non-text nodes (i.e. Table, Figure, etc.) distillation. So, we cannot use simple L2 distance \cite{chen2021deep} in order to compute the similarity between nodes and edges during distillation. Hence we used cosine similarity to compute the similarity between the teacher and student nodes as it considers the orthogonality of the representation and Mahalanobis distance for the edges which is sensitive to the outliers. Also, we propose a sample mining approach for the "Text" nodes to strategically remove less important text-associated edges. This not only cuts down the biased edges but also improves the regularization of false negatives in distillation framework.

The overall contributions of this work can be summarized as follows:
\begin{itemize}
    \item A new task knowledge distillation in document object detection has been proposed and solved through a structure instance graph creation whose node contains the RoI instances of the RPN and the edges represent the relationship between the nodes through feature similarity. To the best of our knowledge, this is the first time where KD for DOD has been explored.
    \item A new sample mining technique has been introduced to get rid of the text bias and improve the regularization of false negatives during distillation.
    \item  Last but not the least, we utilize cosine similarity for the node-to-node distillation to tackle the orthogonality and Mahalanobis distance for edge-to-edge distillation to tackle the outliers which allows us to perform heterogeneous distillation which is one of the most critical problems in distillation till date.
\end{itemize}

The rest of the paper is organized in the following way: In \cref{s:sota} we review state-of-the-art approaches (SOTA) for DOD and KD. We describe the \emph{GraphKD} in \cref{s:method}. We introduce our experimental evaluation as well as ablation studies in \cref{s:experiments} and discuss the extensive experimentation to consolidate our claims. Finally, \cref{s:conclusion} concludes and guides the future research directions.

%% file: tex/sota.tex
Mainstream DOD algorithms were initially dominated by the classical heuristic rule-based algorithms before the rise of deep learning. Then, Convolution frameworks took the lead until transformers-based architectures demonstrated remarkable performance. This section provides a brief overview of SOTA methodologies for DOD till date. As there is no work on KD for DOD, we are trying to introduce a generalized SOTA on KD for object detection.

\paragraph{\textbf{Heuristic Rule-based DOD:}}
This refers to a method of identifying and extracting structured information from documents using a set of predefined rules and heuristics. Heuristics methods can be further classified into top-down, bottom-up, and hybrid approaches based on the parsing directions. Top-down strategies \cite{journet2005text,kise1998segmentation} perform iteratively partitioning a document image into regions until a distinct region is identified. This strategy offers quicker implementation but sacrifices generalization and is effective primarily on specific document types. On the other hand, Bottom-up approaches \cite{asi2015simplifying,saabni2011language} perform pixel grouping, merging, and other set operations to create homogeneous regions around similar objects, separating them from dissimilar ones. Although bottom-up approaches can tackle complex layouts, they are computationally expensive, especially for large or high-resolution documents. Moreover, to take advantage of both, hybrid strategies \cite{chen2011table,fang2011table} leverage both bottom-up and top-down cues for fast and efficient results. While effective for table detection in the pre-deep learning era, these methods fall short for other complex categories.

\paragraph{\textbf{Convolution-based DOD:}}
the use of convolutional neural networks (CNNs) leverages the power of convolutional operations to learn hierarchical features and spatial relationships in the document images, enabling the network to effectively distinguish between different types of document components. In 2015, Faster-RCNN \cite{sun2019faster} provides a strong baseline for table detection and further extended to solve page segmentation \cite{li2020page}. With an extra segmentation loss and additional branch on top of Faster-RCNN for mask prediction, Mask-RCNN \cite{li2020page} provides the instance segmentation benchmark for newspaper elements. Similarly, RetinaNet \cite{lin2021keyword} provides a complex convolution benchmark for keyword detection in document images, specifically focusing on text region detection. Meanwhile, DeepDeSRT \cite{schreiber2017deepdesrt} introduces a novel image transformation strategy to set a new benchmark for table detection and structure recognition. It identifies visual features of table structures and feeds them into a fully convolutional network with skip pooling. A similar FCNN-based framework \cite{oliveira2018dhsegment,rahal2023layout} has been utilized for historical documents which surpasses the previous convolutional auto-encoder benchmark with transfer learning paradigms \cite{chen2015page,chen2017convolutional,saha2019graphical} on ICDAR2017 Page Object Detection (POD) dataset. A new cross-domain DOD benchmark was established in \cite{li2020cross} to apply domain adaptation strategies to solve the domain shift problem with DeepLabv3+ \cite{markewich2022segmentation} and YOLO \cite{deng2023yolo}. However, the problem isn't fully solved until the arrival of transformers due to the lack of cross-attention mechanism. Lastly, a new benchmark \cite{yang2021vision} for vision-based layout detection utilized a recurrent convolutional neural network with VoVNet-v2 backbone, generating synthetic PDF documents from ICDAR-2013 and GROTOAP datasets, achieving a new benchmark for scientific document segmentation.

\paragraph{\textbf{Transformer based DOD:}}
Nowadays, modern transformers excel in the performance of DOD by using positional embedding and self-attention mechanisms \cite{liao2023doctr}. Here, DiT \cite{li2022dit} set a new baseline for DOD, using self-supervised pretraining on large-scale unlabeled document images, but its applicability to small magazine datasets like PRIMA is limited. To enhance performance, TILT \cite{powalski2021going} introduces a mechanism that concurrently learns textual semantics, visual features, and layout information using an encoder-decoder Transformer. A similar auto-encoder mechanism \cite{yang2022transformer} establishes a new baseline for the PubLayNet dataset (AP: 95.95) by extracting text information through OCR. Recently, LayoutLmv3 \cite{huang2022layoutlmv3} achieved state-of-the-art results in visual document understanding tasks through joint learning of text, layout, and visual features. While excelling for large-scale datasets, it also falls short for small-scale datasets like DiT. Other recent approaches \cite{da2023vision,zhong2023hybrid,tang2023unifying,douzon2023long,zhang2021vsr} leverage joint pretraining for various VDU tasks, including document visual question answering. While beneficial for downstream tasks through unified pretraining, they exhibit a pretraining bias, hindering performance in domain shifts and struggling to learn class information with low instance numbers due to a lack of weight prioritization. SwinDocSegmenter \cite{banerjee2023swindocsegmenter} tries to solve this problem through contrastive denoising training and hybrid bipartite matching, however, it is computationally expensive (223M model parameters to train). SemiDocSeg \cite{banerjee2023semidocseg} tries to improve the training strategy by training only the categories that occur rarely in the document and letting the other categories learn through the co-occurrence matrix and support set. It improves the training time of the system and reduces the annotation time but it still has 223M parameters to train. This highlights the necessity of a KD technique for optimizing the number of trainable parameters.

\paragraph{\textbf{Knowledge Distillation:}}
According to our best knowledge, there is no such work on knowledge distillation for document object detection. So, we are striving to achieve the closest state-of-the-art results available for knowledge distillation in document object detection. KD strategies can be categorized into three main categories: \textit{response-based} KD~\cite{aditya2019spatial,ba2014deep,hinton2015distilling,mirzadeh2020improved,yang2023knowledge,Zhao_2022_CVPR} seeks to match the final layer predictions of the teacher model; \textit{feature-based} KD~\cite{ahn2019variational,chen2021cross,chen2017learning,chen2021distilling,heo2019knowledge,komodakis2017paying} aims to mimic features extracted from intermediate hidden layers of the deep network and \textit{relation-based} KD~\cite{chawla2021data,kang2021instance,zhang2020improve,dai2021general} which exploits the relations between different layers or sampled data points. However, the latter approach is more geared toward pixel-based semantic segmentation tasks. While feature-based KD is more versatile, it is more expensive and harder to implement than soft teacher predictions. While offline methods~\cite{de2022structural,li2023object} consider an existing frozen teacher model, online methods~\cite{zhang2023structured,zheng2022localization} update both student and teacher networks jointly. Self-distillation~\cite{wu2022single,wu2023spatial} represents a special case of online KD, which employs the same network as both the teacher and student, progressively outperforming the network's performance, albeit disregarding the aim of efficiency.

However, all the methods described above, perform layerwise distillation so the two backbones should follow the same architecture (i.e. CNN to CNN, ViT to ViT). There is no chance of heterogeneous distillation (i.e. ViT to CNN or vice-versa). This problem motivates us to propose graph-based knowledge distillation (GraphKD) which gives us the freedom about the choice of backbone. In fact, we can distill an entire teacher model to a simple student backbone, as it performs node-to-node as well as edge-to-edge distillation via a distillation loss.

%% file: tex/method.tex
\begin{figure}[!htbp]
\centering
  \includegraphics[width=\linewidth]{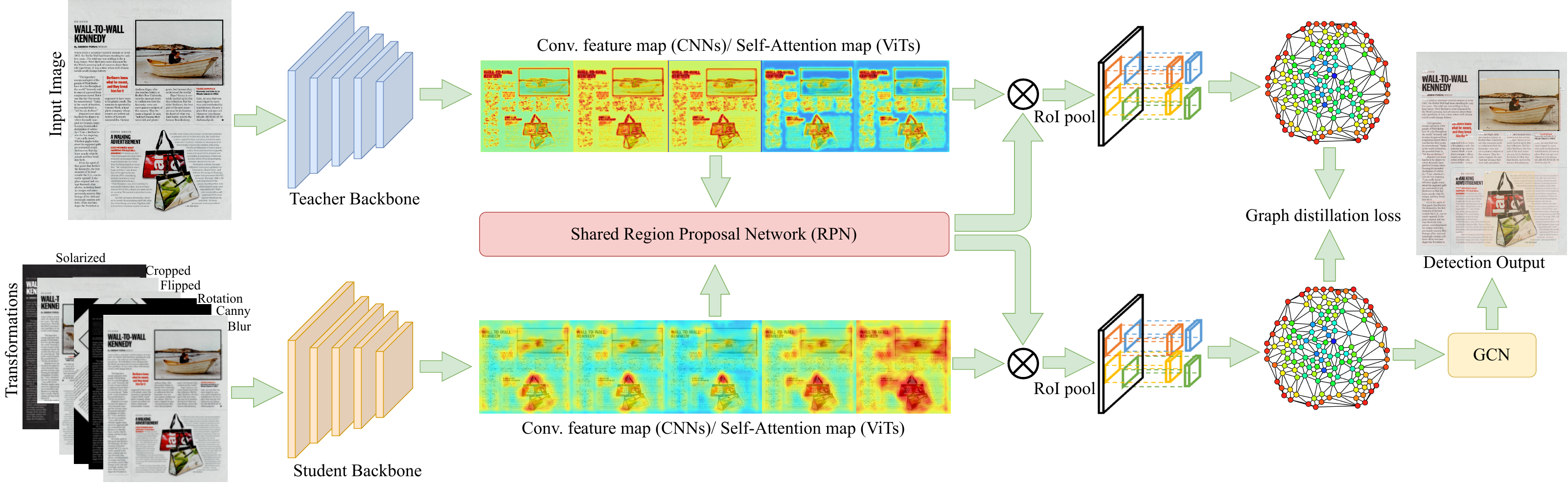}
\caption{\textbf{GraphKD} creates a graph from RoI pooled features of both teacher and student networks and utilize graph distillation loss from knowledge transfer. Finally, a Graph Convolution Network has been used to predict the object classes.}
\label{fig:graph}
\vspace{-4mm}
\end{figure}

We introduce a distillation framework  in \cref{fig:graph} where we are trying to generate a structured instance graph constructed using regional objects within both the teacher and student backbone. This graph effectively leverages the profound insights embedded within detection networks, essentially representing a knowledge framework integrated into the detection system. Distilling this structured graph not only facilitates thorough knowledge transfer but also preserves the entirety of the topological structure corresponding to its embedding space.
It provides us the flexibility to perform knowledge distillation from any large network to a smaller network (e.g. ResNet18) without depending on its architectural similarity as we are performing a node-to-node distillation between two structured graph instances which is a main drawback of the SOTA distillation strategy. 

\subsection{Structured Instance Graph Construction}
\begin{figure}[!htbp]
\vspace{-4mm}
\centering
  \includegraphics[width=\textwidth]{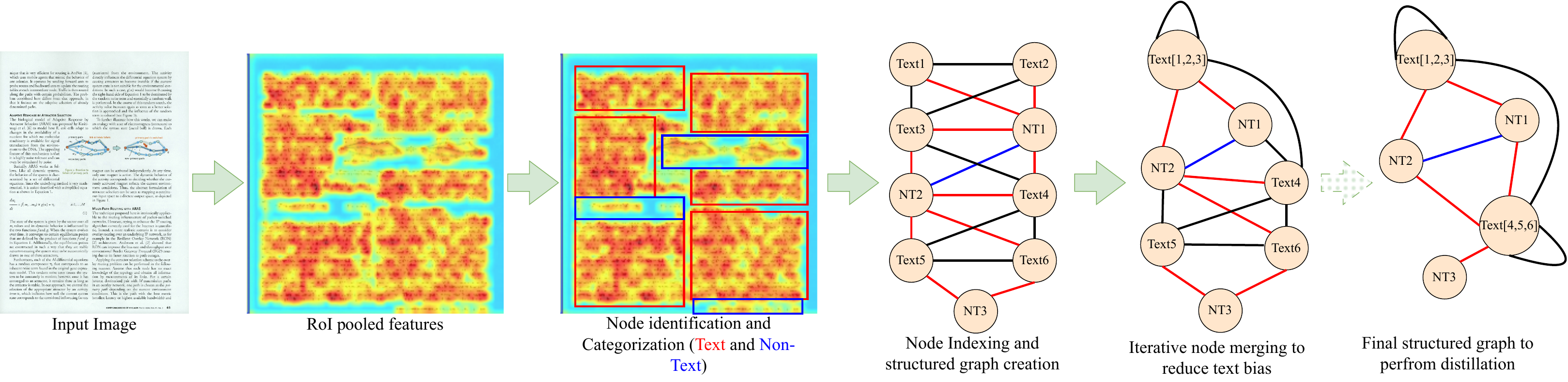}
\caption{\textbf{Structured graph creation:} Here first we extracted the RoI pooled features and classified them into \textcolor{red}{"Text"} and \textcolor{blue}{"Non-text"} based on their covariance. Then we initialize the node in the identified RoI regions and define the adjacency edges. Lastly, we iteratively merge the text node with an adaptive sample mining strategy to reduce text bias.}
\label{fig:sgc}       
\vspace{-4mm}
\end{figure}

Within the structured graph, each node is a representation of an individual instance within an image, and this representation is in the form of a vectorized feature associated with that specific instance. Also, the connection between two instances is established as the edge connecting their respective nodes, and it is determined by measuring their similarity within the embedding space. Indeed, it's essential to highlight that the definition and meaning of an edge in this context are fundamentally distinct from the concept of pixel similarity as outlined in \cite{chen2021deep,liu2019structured}. It is notable that, unlike the other approaches that handle the entire backbone feature map, our focus is directed toward constructing graphs using Region of Interest (RoI) pooled features (See 2nd and 3rd diagram of  \cref{fig:sgc}). These RoI features are derived from the RPN proposals and are then transmitted to the subsequent detection head for further processing. Our nodes are obtained through pooling and extraction using adaptable semantic proposals of different scales and sizes, thereby fostering robust semantic connections among them while those \cite{chen2021deep,liu2019structured} re-sampled and uniformly distributed pixel blocks with the same sizes within an image. The robust connections established between instances that are transferred from the teacher to the student play a pivotal role in enabling interpretable distillation within our approach. Along with that, this framework shared the teacher and student ROIs to make a perfect alignment of the anchor boxes so the same regional features get highlighted, and extracted by two different backbones (teacher and student).

\subsection{Node Definition}
We create nodes directly from RoI pooled features and categorize them into "non-text" or "text" based on the covariance of the RoI pooled features. As depicted in the \cref{fig:sgc} (2nd diagram), the "text" region contains high feature covariance compared to the "non-text" nodes. However, as depicted in the \cref{fig:sgc}, there are 6 "text" nodes and 3 "non-text" nodes, which leads to a large bias towards the text region in the final object detection performance (i.e. lots of regions will be misclassified as text region).  In order to reduce this text bias, we concatenate the adjacency text nodes whose edge distance is below a certain threshold (i.e. node merging and edge reduction). In the late stage, when we need to perform the node separation to detect individual text regions we utilize an adaptive text weight loss, which performs a weight base separation until maximum IoU with the ground truth is achieved.

\subsection{Edge Definition}
The edges of the structured graph $\mathcal{G}_s$ as $\xi_s = [e_{ij}]_{n \times n}$ where $n$ is the size of the node feature set. $e_{pq}$ is the edge between p-th and q-th nodes, denoting the cosine similarity of the corresponding instances in the graph embedding space as defined in \cref{eq5}.
\begin{equation}
    \label{eq5}
    e_{pq} = \dfrac{v_p \cdot v_q}{||v_p|| \cdot ||v_q||}
\end{equation}

Where, $v_i$ is the node of the $i^{th}$ instance. This definition of the edges is invariable to the length of the feature belonging to the corresponding node. This helps the framework to get rid of backbone similarity. The graph we are creating is always a complete, symmetric, and undirected graph (i.e. $e_{pq} = e_{qp}$) with elements all being 1 in the principal diagonal. We utilize this cosine similarity to create the weighted edges because it helps to put small weights between the edges of the similar nodes (i.e. text-text, \textcolor{blue}{non-text(NT)-non-text(NT)}) and higher weights between the edges of the different nodes (\textcolor{red}{text-NT}).  This helps a lot in thresholding-based node merging and edge classification in the later stage.

\subsection{Node Indexing}
We've found that using dense edges generated by the entire set of nodes can hinder the training process (See 4th diagram of \cref{fig:sgc}). This is because a significant number of "Text" nodes contribute to an excessive loss during the distillation of object-related edges. To counteract this issue, it's beneficial to create a more focused set of edges using only the non-text nodes.

However, eliminating all edges related to "Text" nodes can lead to a significant loss of information early in the training process. This is because some of these edges represent hard negative samples that provide valuable insights during training. To address this, we've introduced a technique called Object Samples Mining \cite{mooney2005mining}. This method identifies and selects pertinent non-text nodes, which are then combined with all the merged "Text" nodes to form edges.
\vspace{-4mm}
\begin{algorithm}
\centering
\caption{Node Indexing}\label{algo1}
\begin{algorithmic}
\Require $V_T^{text}, V_S^{text}, T$ \Comment{$T$: Teacher model; $S$: Student Model;}
\Ensure $V_T^{mine}, V_S^{mine}$ \Comment{$mine$:Nodes after mining}
\State Initialize threshold $t$.
\State $\mathcal{L}_{RoI_T^{text}} \leftarrow T.RoIclsLoss(V_T^{text})$
\State $V_{ind} \leftarrow V_T^{text}: \forall V_T^{text} \models {L}_{RoI_T^{text}} \gg t$
\State $V_T^{mine} \leftarrow SelectIndex(V_T^{bg},V_{text})$
\State $V_S^{mine} \leftarrow SelectIndex(V_S^{bg},V_{text})$
\State \textbf{Return} $V_T^{mine}, V_S^{mine}$
\end{algorithmic}
\end{algorithm}
\vspace{-4mm}

\cref{algo1} presents a method that merges specific "Text" samples based on a criterion: their classification losses in the teacher model exceed a threshold \textit{t} (empirically decided). This suggests that these "Text" samples are more likely to be misclassified (i.e. leads to text bias). Consequently, they can be appropriately included in the set of edges that only contain non-text nodes. This inclusion ensures that a dense graph is maintained.

In this \cref{algo1}, samples with high confidence to be classified to the "Text" are not directly added to the set. Here, we perform node indexing through the $SelectIndex$ function as defined in \cite{shrivastava2016training}. After this sample mining, we applied graph distillation loss to perform KD from the teacher-to-student model.

\subsection{Graph Distillation Loss}
The graph distillation loss $\mathcal{L}_g$ is defined as the variance between structured graphs of teacher $\mathcal{G}_{s,t}$ and student $\mathcal{G}_{s,s}$, consisting of graph node and edge loss ($\mathcal{L}_v$, and $\mathcal{L}_\xi$) respectively. We simply compute the Mahalanobis distance function to evaluate this loss as depicted in \cref{eq6}.
\begin{equation}
\begin{split}
    \label{eq6}
    \mathcal{L}_g = \lambda_1 \cdot \mathcal{L}_v^{text} + \lambda_2 \cdot \mathcal{L}_v^{nt} + \lambda_3 \cdot \mathcal{L}_\xi
    = \dfrac{\lambda_1}{N_{nt}}\sum_{i=1}^{N_{nt}} ||\dfrac{v_i^{t,nt} - v_i^{s,nt}}{\sigma_{nt}}|| + \\
    \dfrac{\lambda_2}{N_{text}}\sum_{i=1}^{N_{text}} ||\dfrac{v_i^{t,text} - v_i^{s,text}}{\sigma_{text}}|| + \dfrac{\lambda_3}{N^2}\sum_{i=1}^{N}\sum_{j=1}^{N}||\dfrac{e_{ij}^t - e_{ij}^s}{\sigma_\xi}||
\end{split}
\end{equation}

where, $\lambda_i$ represents the penalty coefficient in order to balance the text, non-text, and edge components of graph distillation loss. We set $\lambda_1$ and $\lambda_3$ to 0.3 based on the Optuna \cite{akiba2019optuna} search, and $\lambda_2 = \alpha \cdot \dfrac{N_{nt}}{N_{text}}$ as an adaptive loss weight for text nodes to mitigate the imbalance problem, where $\alpha$ is empirically determined to ensure that the loss is on a similar scale as other distillation losses.

The graph node loss $\mathcal{L}_v$ is the imitation loss \cite{negrinho2018learning} between node set which makes an accurate alignment between teacher and student backbone extracted features through instance matching (See \cref{algo2}). Conventionally, in KD, there's a preference for matching the feature map directly between two networks. But when it comes to detection models, not every pixel in the feature maps contributes to the classification and box regression loss calculation. Instead of using the entire feature map, we use selected features from both text and non-text regions to compute the graph node loss. This method helps the student model to pay attention to the RoIs and assimilate relevant object knowledge.
\vspace{-4mm}
\begin{algorithm}
\centering
\caption{Graph Distillation Loss}\label{algo2}
\begin{algorithmic}
\Require Image $x$, transformations $x_t$, teacher $T$, student $S$
\Ensure $\mathcal{L}_g$
\State $\mathcal{F}_T \leftarrow$ $T$.backbone$(x)$
\State $\mathcal{F}_s \leftarrow$ $S$.backbone$(x_t)$
\State proposals $\leftarrow S$.rpn($x, x_t, \mathcal{F}_s$)
\State $V_T^{nt}, V_T^{text} \leftarrow$ RoIPool($\mathcal{F}_T$, proposals)
\State $V_S^{nt}, V_S^{text} \leftarrow$ RoIPool($\mathcal{F}_S$, proposals)
\State $V_T^{mine}, V_S^{mine} \leftarrow$ Node Indexing($V_T^{text}, V_S^{text}, T$)\
\State $V_T \leftarrow  V_T^{nt} \oplus V_T^{mine}$
\State $V_S \leftarrow  V_S^{nt} \oplus V_S^{mine}$
\State $\xi_T \leftarrow$ CosineSimilarity($V_T$)
\State $\xi_S \leftarrow$ CosineSimilarity($V_S$)
\State $\mathcal{L}_v^{text} \leftarrow$ MahalanobisDistance($V_T^{Text},V_S^{Text}$)\
\State $\mathcal{L}_v^{nt} \leftarrow$ MahalanobisDistance($V_T^{nt},V_S^{nt}$)
\State $\mathcal{L}_\xi \leftarrow$ MahalanobisDistance($\xi_T,\xi_S$)
\State \textbf{Return} $\mathcal{L}_v^{text} + \mathcal{L}_v^{nt} + \mathcal{L}_\xi$
\end{algorithmic}
\end{algorithm}
\vspace{-4mm}

As depicted in \cref{algo2} $\mathcal{L}_\xi$ is also an imitation loss between the entire edge set. This helps in aligning the edges of the student with those of the teacher. In our tests, imitating features doesn't fully tap into the depth of available knowledge. If the node loss doesn't capture the intricate semantic relationships effectively, then the edge loss steps in to drive the learning of these pairwise interactions. Hence, to synchronize the knowledge structure between the student and teacher, it's vital to craft the edge loss in a way that encapsulates the overarching structured data within detectors. The only thing left is to match the output logits. We utilize a KL Divergence loss in order to incorporate these logits matching as defined in \cite{yang2021learning}.

After this stage, the student model can correctly identify object regions (i.e. the bounding boxes) in a document. However, it doesn't have any about object categories except "Text" and "non-text" labels. In order to, further categorize the "non-text node" and predict the correct object labels we utilize a GNN \cite{wang2020joint} with a cross-entropy loss for node classification.  

%% file: tex/experiments.tex
For validation purposes, we have considered four important benchmark datasets (PubLayNet \cite{zhong2019publaynet}, PRIMA \cite{clausner2019icdar2019}, Historical Japanese \cite{shen2020large}, and DoclayNet \cite{pfitzmann2022doclaynet}) which cover most of the existing document object categories (see \cref{tab:01}). Our experimentation shows that the proposed GraphKD provides similar results to the large-scale supervised models (SwinDocSegmenter \cite{banerjee2023swindocsegmenter}, LayoutLMv3 \cite{huang2022layoutlmv3}, DocSegTr \cite{biswas2022docsegtr}, and so on) with a lesser no. of parameters.

\begin{table}[!htbp]
\vspace{-4mm}
\caption{Experimental dataset description (instance level)}
\label{tab:01}
\centering
\resizebox{\textwidth}{!}{
\begin{tabular}{@{}cccccccccccc@{}}
\toprule
\multicolumn{3}{c}{\textbf{PubLayNet}}                                                                           & \multicolumn{3}{c}{\textbf{PRIMA}}                                                                              & \multicolumn{3}{c}{\textbf{Historical Japanese}}                                                                & {\textbf{DocLayNet}}                                                                          \\ \midrule
\multicolumn{1}{c}{\textbf{Object}} & \multicolumn{1}{c}{\textbf{Train}} & \multicolumn{1}{c}{\textbf{Eval}} & \multicolumn{1}{c}{\textbf{Object}} & \multicolumn{1}{c}{\textbf{Train}} & \multicolumn{1}{c}{\textbf{Eval}} & \multicolumn{1}{c}{\textbf{Object}} & \multicolumn{1}{c}{\textbf{Train}} & \multicolumn{1}{c}{\textbf{Eval}}  & \multicolumn{1}{c}{\textbf{Object}} & \multicolumn{1}{c}{\textbf{Train}} & \multicolumn{1}{c}{\textbf{Eval}} \\ \midrule
\multicolumn{1}{c}{Text}            & \multicolumn{1}{c}{2,343,356}      & \multicolumn{1}{c}{88,625}        & \multicolumn{1}{c}{Text}            & \multicolumn{1}{c}{6401}           & \multicolumn{1}{c}{1531}          & \multicolumn{1}{c}{Body}            & \multicolumn{1}{c}{1443}           & \multicolumn{1}{c}{308}                   & \multicolumn{1}{c}{Caption}         & \multicolumn{1}{c}{20280}          & \multicolumn{1}{c}{1543}          \\ \midrule
\multicolumn{1}{c}{Title}           & \multicolumn{1}{c}{627,125}        & \multicolumn{1}{c}{18,801}        & \multicolumn{1}{c}{Image}           & \multicolumn{1}{c}{761}            & \multicolumn{1}{c}{163}           & \multicolumn{1}{c}{Row}             & \multicolumn{1}{c}{7742}           & \multicolumn{1}{c}{1538}                      & \multicolumn{1}{c}{Footnote}        & \multicolumn{1}{c}{5964}           & \multicolumn{1}{c}{387}           \\ \midrule
\multicolumn{1}{c}{Lists}           & \multicolumn{1}{c}{80,759}         & \multicolumn{1}{c}{4239}          & \multicolumn{1}{c}{Table}           & \multicolumn{1}{c}{37}             & \multicolumn{1}{c}{10}            & \multicolumn{1}{c}{Title}           & \multicolumn{1}{c}{33,637}         & \multicolumn{1}{c}{7271}                      & \multicolumn{1}{c}{Formula}         & \multicolumn{1}{c}{22367}          & \multicolumn{1}{c}{1966}          \\ \midrule
\multicolumn{1}{c}{Figures}         & \multicolumn{1}{c}{109,292}        & \multicolumn{1}{c}{4327}          & \multicolumn{1}{c}{Math}            & \multicolumn{1}{c}{35}             & \multicolumn{1}{c}{7}             & \multicolumn{1}{c}{Bio}             & \multicolumn{1}{c}{38,034}         & \multicolumn{1}{c}{8207}                       & \multicolumn{1}{c}{List-item}       & \multicolumn{1}{c}{170889}         & \multicolumn{1}{c}{10522}         \\ \midrule
\multicolumn{1}{c}{Tables}          & \multicolumn{1}{c}{102,514}        & \multicolumn{1}{c}{4769}          & \multicolumn{1}{c}{Separator}       & \multicolumn{1}{c}{748}            & \multicolumn{1}{c}{155}           & \multicolumn{1}{c}{Name}            & \multicolumn{1}{c}{66,515}         & \multicolumn{1}{c}{7257}                       & \multicolumn{1}{c}{Page-footer}     & \multicolumn{1}{c}{64717}          & \multicolumn{1}{c}{3994}          \\ \midrule
\multicolumn{1}{c}{-}               & \multicolumn{1}{c}{-}              & \multicolumn{1}{c}{-}             & \multicolumn{1}{c}{other}           & \multicolumn{1}{c}{86}             & \multicolumn{1}{c}{25}            & \multicolumn{1}{c}{Position}        & \multicolumn{1}{c}{33,576}         & \multicolumn{1}{c}{7256}                      & \multicolumn{1}{c}{Page-header}     & \multicolumn{1}{c}{50700}          & \multicolumn{1}{c}{3366}          \\ \midrule
\multicolumn{1}{c}{-}               & \multicolumn{1}{c}{-}              & \multicolumn{1}{c}{-}             & \multicolumn{1}{c}{-}               & \multicolumn{1}{c}{-}              & \multicolumn{1}{c}{-}             & \multicolumn{1}{c}{Other}           & \multicolumn{1}{c}{103}            & \multicolumn{1}{c}{29}                         & \multicolumn{1}{c}{Picture}         & \multicolumn{1}{c}{39621}          & \multicolumn{1}{c}{3534}          \\ \midrule
\multicolumn{1}{c}{-}               & \multicolumn{1}{c}{-}              & \multicolumn{1}{c}{-}             & \multicolumn{1}{c}{-}               & \multicolumn{1}{c}{-}              & \multicolumn{1}{c}{-}             & \multicolumn{1}{c}{-}               & \multicolumn{1}{c}{-}              & \multicolumn{1}{c}{-}                          & \multicolumn{1}{c}{Section-header}  & \multicolumn{1}{c}{18003}               & \multicolumn{1}{c}{8550}          \\ \midrule
\multicolumn{1}{c}{-}               & \multicolumn{1}{c}{-}              & \multicolumn{1}{c}{-}             & \multicolumn{1}{c}{-}               & \multicolumn{1}{c}{-}              & \multicolumn{1}{c}{-}             & \multicolumn{1}{c}{-}               & \multicolumn{1}{c}{-}              & \multicolumn{1}{c}{-}                          & \multicolumn{1}{c}{Table}           & \multicolumn{1}{c}{30115}               & \multicolumn{1}{c}{2394}          \\ \midrule
\multicolumn{1}{c}{-}               & \multicolumn{1}{c}{-}              & \multicolumn{1}{c}{-}             & \multicolumn{1}{c}{-}               & \multicolumn{1}{c}{-}              & \multicolumn{1}{c}{-}             & \multicolumn{1}{c}{-}               & \multicolumn{1}{c}{-}              & \multicolumn{1}{c}{-}                          & \multicolumn{1}{c}{Text}            & \multicolumn{1}{c}{431222}               & \multicolumn{1}{c}{29940}         \\ \midrule
\multicolumn{1}{c}{-}               & \multicolumn{1}{c}{-}              & \multicolumn{1}{c}{-}             & \multicolumn{1}{c}{-}               & \multicolumn{1}{c}{-}              & \multicolumn{1}{c}{-}             & \multicolumn{1}{c}{-}               & \multicolumn{1}{c}{-}              & \multicolumn{1}{c}{-}                          & \multicolumn{1}{c}{Title}           & \multicolumn{1}{c}{4423}               & \multicolumn{1}{c}{335}           \\ \midrule
\textbf{Total}                        & \textbf{3,263,046}                  & \textbf{120,761}                   & \textbf{Total}                       & \textbf{8068}                       & \textbf{1891}                      & Total                                & 181,097                             & 31,866                                                           & \textbf{Total}                       & \textbf{994123}                     & \textbf{66531}                     \\ \bottomrule
\end{tabular}
}
\end{table}
 It has been observed in \cref{tab:01}, that PRIMA \cite{clausner2019icdar2019} has only 8068 no. of training instances however "Table", "Math", and "Other" region have only 37, 35, and 86 no. of training instances which is not sufficient to define an object class makes the dataset more challenging to solve. On the other hand, PublayNet \cite{zhong2019publaynet} and HJ \cite{shen2020large} have only 5 and 7 object classes respectively, and all the classes have sufficient no. of training instances except the "Others" in HJ. However, it is very difficult to define the "Others" class properly. Last but not the least, DoclayNet \cite{pfitzmann2022doclaynet} has 11 classes and provides a large variability bias towards some classes (i.e. the training samples are not equally distributed. For example, List-item has $\approx 170K$ training instances whereas, Title has only $\approx 4K$ training instances.).

 Now, in order to evaluate the results on the aforementioned dataset, we utilize the IoU score which assesses instance segmentation accuracy, with COCO benchmarks using mean AP across IoU thresholds (0.5 to 0.95 with a step size of 0.05) to calculate mAP. This metric has been used throughout the rest of the paper to evaluate the performance of the proposed settings.
 
\subsection{Ablation Studies}
In this context, We have performed two sets of the ablation. In the first set (see \cref{tab:gr_ab}), we have used the various combinations of components to emphasize the distinct contribution of each element in our graph distillation approach. Our framework incorporates three distinct modules that contribute to the graph distillation loss: 1) edge, 2) text node, and 3) non-text node. It has been observed that maintaining a consistent edge structure between student and teacher models adds $\approx 3\%$  AP points to the distillation outcome. This suggests that beyond direct pixel-to-pixel imitation, simply ensuring alignment in relationships can serve as a critical regularization to maintain topological configuration. This validates the effectiveness of our approach.

The introduction of student features mirroring non-text object nodes leads to a notable increase of $\approx 4\%$ in Average Precision (AP), surpassing the impact achieved by considering edges. This implies that distilling non-text features effectively encourages the student networks to prioritize regions containing non-text instances (i.e. helps to reduce the text bias). This underscores the significance of feature alignment within these foreground-labeled regions as more crucial for the student to emulate than the broader and potentially noisy high-dimensional feature maps. Additionally, when edges collaborate with nodes, it produces even more promising outcomes, affirming the effectiveness of both components of the graph, as they complement each other.

\begin{table}[!htbp]
\vspace{-4mm}
\centering
\caption{Ablation study on the building blocks of GraphKD with DocLayNet using resnet50 to resnet18 setup}
\label{tab:gr_ab}
\begin{tabular}{@{}cccccccccc@{}}
\toprule
Student & Edge & Non-text & Text & AP            & AP@50         & AP@75         & APs           & APm           & APl           \\ \midrule
\checkmark       &      &          &      & 33.1          & 52.5          & 36.0          & 19.1          & 36.6          & 41.3          \\
\checkmark       & \checkmark    &          &      & 36.6          & 59.8          & \textbf{40.1} & 21.2          & \textbf{38.1} & 41.4          \\
\checkmark       & \checkmark    & \checkmark        &      & 40.2          & 62.1          & 33.8          & 24.7          & 34.2          & 41.2          \\
\checkmark       & \checkmark    & \checkmark        & \checkmark    & \textbf{42.1} & \textbf{80.5} & 36.3          & \textbf{29.0} & 34.7          & \textbf{41.8} \\ \midrule
\multicolumn{4}{c}{Teacher}      & 61.2          & 87.9          & 46.3          & 30.1          & 40.2          & 42.7          \\ \bottomrule
\end{tabular}
\vspace{-4mm}
\end{table}

Lastly, incorporating the imitation of student features within text regions results in an additional increase in AP, specifically a gain of $\approx 2\%$ when compared to the gains achieved with foreground nodes. This indicates that, even when the primary focus is on imitating non-text object nodes, the inclusion of text nodes becomes significant in the process of student distillation, particularly when balanced using our adaptive Mahalanobis distance loss. (NOTE: in \cref{tab:gr_ab} there is a large performance gap between the performance of teacher and student models as ResNet50 and ResNet18 pose different numbers of layers in each of the ResNet blocks, so some important feature has been lost during the feature compression in node-to-node distillation.)

In the second set, we have performed the ablation of different distance functions we can use to perform the node-to-node and edge-to-edge knowledge distillation (see \cref{tab:gr_ab1}). It has been observed that, L1 and L2 distance as a loss function performs the worst as it only computes the absolute error or least square errors between the nodes or the edges. It neither considers the similarity nor their covariance. So when we are distilling using L1 or L2 losses it just penalizes the error based on the feature values of the node or the weights of the edges.

\begin{table}[!htbp]
\vspace{-4mm}
\centering
\caption{Ablation study on the loss functions combinations of GraphKD with DocLayNet using resnet50 to resnet18 setup}
\label{tab:gr_ab1}
\begin{tabular}{@{}cccccccc@{}}
\toprule
Node2Node           & Edge2Edge   & AP   & AP@50 & AP@75 & APs  & APm  & APl  \\ \midrule
\multirow{4}{*}{L1} & L1          & 3.0  & 6.1   & 2.8   & 0.4  & 1.6  & 2.2  \\
                    & L2          & 7.1  & 8.3   & 3.5   & 1.2  & 1.6  & 4.3  \\
                    & Cosine      & 7.1  & 10.5  & 8.2   & 1.5  & 2.1  & 6.7  \\
                    & Mahalanobis & 8.9  & 11.2  & 9.0   & 4.4  & 3.3  & 10.2 \\ \midrule
\multirow{4}{*}{L2} & L1          & 10.2 & 11.9  & 14.6  & 6.0  & 7.5  & 12.7 \\
                    & L2          & 10.2 & 13.4  & 14.8  & 6.8  & 7.8  & 20.7 \\
                    & Cosine      & 10.2 & 15.1  & 16.3  & 7.3  & 9.9  & 23.2 \\
                    & Mahalanobis & 12.9 & 18.8  & 23.2  & 10.0 & 12.0 & 24.9 \\ \midrule
\multirow{4}{*}{Mahalanobis} & L1          & 19.5          & 24.8          & 25.6          & 13.1          & 15.4          & 25.5          \\
                    & L2          & 23.5 & 61.7  & 29.4  & 14.4 & 23.0 & 28.1 \\
                    & Cosine      & 24.3 & 63.6  & 29.7  & 20.6 & 23.6 & 30.6 \\
                    & Mahalanobis & 28.5 & 64.5  & 31.6  & 21.9 & 23.6 & 31.7 \\ \midrule
\multirow{4}{*}{Cosine}      & L1          & 30.4          & 66.8          & 31.7          & 23.3          & 26.9          & 36.9          \\
                    & L2          & 38.9 & 68.5  & 32.7  & 25.6 & 28.3 & 37.2 \\
                    & Cosine      & 38.9 & 79.5  & 33.5  & 27.5 & 34.1 & 41.5 \\
                             & Mahalanobis & \textbf{42.1} & \textbf{80.5} & \textbf{36.3} & \textbf{29.0} & \textbf{34.7} & \textbf{41.8} \\ \bottomrule
\end{tabular}
\vspace{-4mm}
\end{table}

On the other hand, when we are using the Cosine distance as a loss function it considers the similarity which is beneficial for node-to-node distillation but not for edge-to-edge distillation. Similarly, Mahalanobis distance considers covariance but not similarity which is also beneficial for edge-to-edge distance. That's why the best combination has been obtained by using Cosine distance for node-to-node distillation and Mahalanobis distance for edge-to-edge distillation.

\subsection{Quantitative Evaluation}
In order to establish a robust quantitative evaluation of GraphKD, we obtain Homogeneous (ResNet152-ResNet101 and ResNet101-Resnet50) as well as Heterogeneous (ResNet50-ResNet18, Resnet101-EfficientnetB0, and ResNet50-MobileNetv2) knowledge distillation across all four competitive benchmarks (PublayNet, PRIMA, Historical Japanese, and DoclayNet respectively). In homogeneous distillation, the number of layers in every ResNet block is the same for the teacher and the student network only we reduce the number of ResNet blocks from teacher to student. On the other hand, in heterogeneous distillation, both the number of layers in every Resnet block and the number of ResNet blocks have been reduced from teacher to student which leads to poor performance compared to homogeneous distillation (which needs only block-wise feature compression) due to double compression of RoI pooled features during knowledge distillation.
\begin{table}[!htbp]
\vspace{-8mm}
\centering
\caption{Graph-based knowledge distillation on PublayNet dataset}
\label{tab:gr_pub}
\resizebox{\textwidth}{!}{
\begin{tabular}{@{}cccccccccccccc@{}}
\toprule
\textbf{} &
  \textbf{\#params (t)} &
  \textbf{\#params (s)} &
  \textbf{Text} &
  \textbf{Title} &
  \textbf{List} &
  \textbf{Table} &
  \textbf{Figure} &
  \textbf{AP} &
  \textbf{AP@50} &
  \textbf{AP@75} &
  \textbf{APs} &
  \textbf{APm} &
  \textbf{APl} \\ \midrule
\textbf{R50-R18}   & 25.6M & 11.1  & 18.9 & 32.2 & 32.9 & 28.9 & 27.0 & 28.0 & 72.8 & 6.7  & 13.7 & 27.0 & 28.6 \\
\textbf{R101-R50}  & 44.5M & 25.6M & 91.0 & 82.9 & 85.2 & 95.0 & 88.9 & 88.6 & 97.0 & 94.5 & 38.6 & 73.8 & 93.2 \\
\textbf{R152-R101} & 60.2M & 44.5M & 90.9 & 82.3 & 85.6 & 95.3 & 89.1 & 88.8 & 97.0 & 94.7 & 38.6 & 73.9 & 93.7 \\
\textbf{R101-EB0}  & 44.5M & 5.3M  & 19.2 & 34.8 & 28.1 & 29.1 & 26.8 & 27.6 & 72.2 & 7.1  & 13.0 & 25.0 & 28.1 \\
\textbf{R50-MNv2}  & 25.6M & 3.4M  & 18.5 & 32.1 & 32.9 & 29.3 & 28.2 & 28.2 & 72.8 & 7.2  & 13.5 & 24.2 & 29.6 \\ \midrule
\textbf{DocSegTr \cite{biswas2022docsegtr}} & - & 168M & 91.1 & 75.6 & 91.5 & 97.9 & 97.1 & 90.4 & 97.9 & 95.8 & - & - & - \\
\textbf{LayoutLMv3 \cite{huang2022layoutlmv3}} & - & 368M & 94.5 & 90.6 & 95.5 & 97.9 & 97.9 & 95.1 & - & -& - & - & - \\
\textbf{SwinDocSegmenter \cite{banerjee2023swindocsegmenter}} & - & 223M & 94.5 & 87.1 & 93.0 & 97.9 & 97.2 & 93.7 & 97.9 & 96.2 & - & - & - \\
\bottomrule
\end{tabular}
}
\vspace{-4mm}
\end{table}

Moreover, in \cref{tab:gr_pub} we have obtained our quantitative evaluation on the PublayNet dataset. For a fair comparison, we compared the performance of our distilled network with state-of-the-art supervised approaches \cite{biswas2022docsegtr,banerjee2023swindocsegmenter,huang2022layoutlmv3} and it is obvious that we cannot outperform those transformer based large scale networks however, we can reduce the performance gap and increase the efficiency. For example, the performance gap between LayoutLMv3 \cite{huang2022layoutlmv3} and distilled ResNet101 Faster-RCNN network is $\approx 7\%$ (Overall AP) however, we can reduce 323.5M(368M-44.5M) number of model parameters. Also, it outperforms DocSegTr (168M) \cite{biswas2022docsegtr} for the "Title" class by $\approx 8\%$ as the Transformers are not the best choice for small object detection.
\begin{table}[!htbp]
\vspace{-8mm}
\centering
\caption{Graph-based knowledge distillation on PRIMA dataset}
\label{tab:gr_p}
\resizebox{\textwidth}{!}{
\begin{tabular}{@{}ccccccccccccccc@{}}
\toprule
 &
  \textbf{\#params(t)} &
  \textbf{\#params(s)} &
  \textbf{Text} &
  \textbf{Image} &
  \textbf{Table} &
  \textbf{Math} &
  \textbf{Separator} &
  \textbf{Other} &
  \textbf{AP} &
  \textbf{AP@50} &
  \textbf{AP@75} &
  \textbf{APs} &
  \textbf{APm} &
  \textbf{APl} \\ \midrule
\textbf{R50-R18}   & 25.6M & 11.1M & 38.5    & 47.7    & 41.6    & 8.5    & 17.5    & 5.4   & 26.5    & 52.1    & 22.3    & 30.8    & 30.9    & 28.3    \\
\textbf{R101-R50}  & 44.5M & 25.6M & 76.6 & 60.8 & 38.3 & 4.1  & 23.4 & 7.1 & 35.0 & 51.0 & 39.0 & 39.6 & 42.1 & 36.6 \\
\textbf{R152-R101} & 60.2M & 44.5M & 79.9 & 64.6 & 44.0 & 30.2 & 25.9 & 6.7 & 41.9 & 58.9 & 42.0 & 41.4 & 41.7 & 43.5 \\
\textbf{R101-EB0} & 44.5M & 5.3M & 20.3 & 19.0 & 15.6 & 4.4 & 13.1 & 3.2 & 12.6 & 40.7 & 1.69 & 16.4 & 18.7 & 12.2\\
\textbf{R50-MNv2} & 25.6M & 3.4M & 19.5 & 16.6 & 28.0 & 7.1 & 12.2 & 5.9 & 14.9 & 40.2 & 2.5 & 16.1 & 20.1 & 15.2\\ \midrule
\textbf{DocSegTr \cite{biswas2022docsegtr}} & - & 168M & 75.2 & 64.3 & 59.4 & 48.4 & 1.8 & 3.0 & 42.5 & 54.2 & 45.8 & - & - & - \\
\textbf{LayoutLMv3 \cite{huang2022layoutlmv3}} & - & 368M & 70.8 & 50.1 & 42.5 & 46.5 & 9.6 & 17.4 & 40.3 & - & -& -&-&- \\
\textbf{SwinDocSegmenter \cite{banerjee2023swindocsegmenter}} & - & 223M & 87.7 & 75.9 & 49.8 & 78.1 & 27.5 & 7.0 & 54.3 & 69.3 & 52.9 & - & - & - \\
\bottomrule
\end{tabular}
}
\vspace{-4mm}
\end{table}

On the other hand, the distilled ResNet101 outperforms LayoutLMv3 \cite{huang2022layoutlmv3} by $\approx 1\%$ which clearly shows large networks are not quite effective for small datasets like PRIMA (see \cref{tab:gr_p}) as in the later stage of training instance they only learn noise due to unavailability of data. Not only that, distilled EfficientnetB0 and MobileNetv2 which have only 5.3M and 3.4M parameters respectively, also outperform the LayoutLMv3 \cite{huang2022layoutlmv3} and DocSegTr \cite{biswas2022docsegtr} for the "Separator class". Similarly, distilled ResNet101 outperforms SwinDocsegmenter \cite{banerjee2023swindocsegmenter} for the "other" classes which shows the real power of the knowledge distillation.

\begin{table}[!htbp]
\vspace{-4mm}
\centering
\caption{Graph-based knowledge distillation on Historical Japanese dataset}
\label{tab:gr_hj}
\resizebox{\textwidth}{!}{
\begin{tabular}{@{}cccccccccccccccc@{}}
\toprule
\textbf{} &
  \textbf{\#params (t)} &
  \textbf{\#params (s)} &
  \textbf{Body} &
  \textbf{Row} &
  \textbf{Title} &
  \textbf{Bio} &
  \textbf{Name} &
  \textbf{Position} &
  \textbf{Other} &
  \textbf{AP} &
  \textbf{AP@50} &
  \textbf{AP@75} &
  \textbf{APs} &
  \textbf{APm} &
  \textbf{APl} \\ \midrule
\textbf{R50-R18}   & 25.6M & 11.1  & 37.6 & 25.9 & 50.1 & 28.0 & 44.0 & 35.7 & 12.4 & 33.4 & 67.2 & 22.7 & 20.7 & 23.3 & 28.9 \\
\textbf{R101-R50}  & 44.5M & 25.6M & 88.2 & 98.3 & 84.9 & 94.6 & 68.6 & 83.5 & 30.2 & 78.3 & 86.0 & 84.7 & 38.2 & 44.6 & 71.5 \\
\textbf{R152-R101} & 60.2M & 44.5M & 98.4 & 98.3 & 85.1 & 94.5 & 68.6 & 84.2 & 28.5 & 79.7 & 87.4 & 86.0 & 37.6 & 43.6 & 74.8 \\
\textbf{R101-EB0}  & 44.5M & 5.3M  & 31.1 & 22.0 & 49.9 & 25.2 & 20.2 & 65.3 & 18.4 & 33.1 & 65.9 & 25.9 & 19.1 & 23.5 & 29.4 \\
\textbf{R50-MNv2}  & 25.6M & 3.4M  & 30.1 & 26.5 & 61.6 & 23.3 & 36.0 & 65.1 & 20.3 & 37.5 & 71.1 & 29.0 & 16.7 & 29.0 & 30.7 \\ \midrule
\textbf{DocSegTr \cite{biswas2022docsegtr}} & - & 168M & 99.0 & 99.1 & 93.2 & 94.7 & 70.3 & 87.4 & 43.7 & 83.1 & 90.1 & 88.1 & - & - & - \\
\textbf{LayoutLMv3 \cite{huang2022layoutlmv3}} & - & 368M & 99.0 & 99.0 & 92.9 & 94.7 & 67.9 & 87.8 & 38.7 & 82.7 & - & - & - & - & - \\
\textbf{SwinDocSegmenter \cite{banerjee2023swindocsegmenter}} & - & 223M & 99.7 & 99.0 & 89.5 & 86.2 & 83.8 & 93.0 & 40.5 & 84.5 & 90.7 & 88.2 & - & - & - \\
\bottomrule
\end{tabular}
}
\vspace{-4mm}
\end{table}

Similarly, for the Historical Japanese dataset (see \cref{tab:gr_hj}) all the supervised methods perform better than the distilled networks as all the classes of this dataset are quite separable from each other. However, an interesting observation has been noticed in the performance of the DoclayNet dataset. Distilled ResNet50 and ResNet101 outperform DocSegTr \cite{biswas2022docsegtr} and LayoutLMv3 \cite{huang2022layoutlmv3} by a significant margin for "Caption", "Page-footer", and "Picture" which shows how the bias factor affects the supervised training and the potential of knowledge distillation to overcome it by forming a more generalized efficient network which we can use in our edge devices.
\begin{table}[!htbp]
\vspace{-4mm}
\centering
\caption{Graph-based knowledge distillation on DocLayNet dataset}
\label{tab:gr_d}
\resizebox{\textwidth}{!}{
\begin{tabular}{@{}cccccccccccccccccccc@{}}
\toprule
 &
  \textbf{\#params(t)} &
  \textbf{\#params(s)} &
  \textbf{Caption} &
  \textbf{Footnote} &
  \textbf{Formula} &
  {\begin{tabular}[c]{@{}c@{}}List \\ item\end{tabular}} &
  {\begin{tabular}[c]{@{}c@{}}Page \\ footer\end{tabular}} &
  {\begin{tabular}[c]{@{}c@{}}Page \\ header\end{tabular}} &
  \textbf{Picture} &
  {\begin{tabular}[c]{@{}c@{}}Section \\ header\end{tabular}} &
  \textbf{Table} &
  \textbf{Text} &
  \textbf{Title} &
  \textbf{AP} \\ \midrule
\textbf{R50-R18}   & 25.6M & 11.1M & 53.4 & 23.3 & 30.6 & 39.7 & 45.7 & 44.2 & 58.4 & 43.0 & 45.6 & 41.5 & 37.3 & 42.1\\
\textbf{R101-R50}  & 44.5M & 25.6M & 77.3 & 46.4 & 48.1 & 72.3 & 60.4 & 63.0 & 72.9 & 59.3 & 73.5 & 77.5 & 64.6 & 65.0\\
\textbf{R152-R101} & 60.2M & 44.5M & 78.9 & 58.1 & 53.7 & 75.3 & 59.3 & 67.3 & 76.4 & 61.6 & 78.0 & 80.3 & 69.0 & 68.9\\
\textbf{R101-EB0} & 44.5M & 5.3M & 36.35 & 20.7 & 27.5 & 32.1 & 26.4 & 34.6 & 24.8 & 32.5 & 25.2 & 25.0 & 32.9 & 28.9\\
\textbf{R50-MNv2} & 25.6M & 3.4M & 28.8 & 13.5 & 20.2 & 24.4 & 26.4 & 22.3 & 25.4 & 25.8 & 24.6 & 24.3 & 23.8 & 23.6\\ \midrule
\textbf{DocSegTr \cite{biswas2022docsegtr}} & - & 168M & 70.1 & 73.7 & 63.5 & 81.0 & 58.9 & 72.0 & 72.0 & 68.4 & 82.2 & 85.4 & 79.9 & 73.4 \\
\textbf{LayoutLMv3 \cite{huang2022layoutlmv3}} & - & 368M & 71.5 & 71.8 & 63.4 & 80.8 & 59.3 & 70.0 & 72.7 & 69.3 & 82.9 & 85.8 & 80.4 & 73.5 \\
\textbf{SwinDocSegmenter \cite{banerjee2023swindocsegmenter}} & - & 223M & 83.5 & 64.8 & 62.3 & 82.3 & 65.1 & 66.3 & 84.7 & 66.5 & 87.4 & 88.2 & 63.2 & 76.8 \\
\bottomrule
\end{tabular}
}
\vspace{-4mm}
\end{table}

\subsection{Comparative Study}
In order to, establish the superiority of the proposed method we have performed a detailed comparative study with a feature-based knowledge distillation method ReviewKD \cite{chen2021distilling}, a logit-based knowledge distillation method NKD \cite{chi2023normkd}, and with a hybrid approach SimKD \cite{chen2022knowledge}. The result of this comparative study has been depicted in \cref{tab:gr_comp}. It has been observed that the performance of the DOD has improved quite a lot through GraphKD than the rest.
\begin{table}[!htbp]
\centering
\caption{Comparative study of GraphKD with the state-of-the-art approaches}
\label{tab:gr_comp}
\resizebox{\textwidth}{!}{
\begin{tabular}{c|c|ccc|ccc|ccc|ccc}
\hline
\multirow{2}{*}{Method} &
  \multirow{2}{*}{Backbone} &
  \multicolumn{3}{c|}{PublayNet} &
  \multicolumn{3}{c|}{PRIMA} &
  \multicolumn{3}{c|}{HJ} &
  \multicolumn{3}{c}{DocLayNet} \\ \cline{3-14} 
                          &           & AP   & AP@50 & AP@75 & AP   & AP@50 & AP@75 & AP   & AP@50 & AP@75 & AP   & AP@50 & AP@75 \\ \hline
\multirow{5}{*}{ReviewKD \cite{chen2021distilling}} & R50-R18   & 20.9 & 56.6  & 3.4   & 17.7 & 29.1  & 12.0  & 19.5 & 36.2  & 19.7  & 18.2 & 60.2  & 21.2  \\
                          & R101-R50  & 72.7 & 94.4  & 90.2  & 22.7 & 38.8  & 26.0  & 62.8 & 79.7  & 74.2  & 61.1 & 84.3  & 74.1  \\
                          & R152-R101 & 77.1 & 95.3  & 90.8  & 26.2 & 44.7  & 30.7  & 66.7 & 77.2  & 69.8  & 63.7 & 84.6  & 75.0  \\
                          & R101-EB0  & 20.1 & 62.5  & 6.7   & 7.1  & 28.6  & 0.9   & 22.1 & 50.9  & 22.8  & 12.7 & 50.1  & 7.6   \\
                          & R50-MNv2  & 19.8 & 63.6  & 6.3   & 9.2  & 20.7  & 1.0   & 25.4 & 58.7  & 15.7  & 14.2 & 44.2  & 5.1   \\ \hline
\multirow{5}{*}{NKD \cite{chi2023normkd}}      & R50-R18   & 17.9 & 29.0  & 1.7   & 9.7  & 10.7  & 8.1   & 27.5 & 61.4  & 20.7  & 14.0 & 58.9  & 16.0  \\
                          & R101-R50  & 68.3 & 89.6  & 81.2  & 12.2 & 28.7  & 16.9  & 52.8 & 59.9  & 74.2  & 37.2 & 57.0  & 40.4  \\
                          & R152-R101 & 70.7 & 90.7  & 82.3  & 16.8 & 34.2  & 20.9  & 56.9 & 65.2  & 69.8  & 41.5 & 62.1  & 45.2  \\
                          & R101-EB0  & 17.7 & 58.0  & 2.7   & 7.0  & 18.7  & 0.7   & 11.7 & 40.8  & 20.3  & 10.2 & 40.1  & 5.6   \\
                          & R50-MNv2  & 13.2 & 63.0  & 3.1   & 8.2  & 10.7  & 0.9   & 15.4 & 38.2  & 17.3  & 12.1 & 34.2  & 5.9   \\ \hline
\multirow{5}{*}{SimKD \cite{chen2022knowledge}}    & R50-R18   & 24.7 & 65.4  & 5.1   & 23.6 & 32.7  & 17.2  & 19.1 & 58.0  & 17.4  & 33.2 & 77.8  & 35.7  \\
                          & R101-R50  & 80.2 & 95.2  & 92.1  & 29.7 & 48.8  & 36.0  & 72.3 & 80.7  & 79.7  & 62.7 & 85.2  & 74.2  \\
                          & R152-R101 & 81.1 & 95.9  & 92.2  & 36.2 & 54.7  & 40.7  & 76.7 & 82.1  & 84.2  & 64.6 & 88.6  & 77.2  \\
                          & R101-EB0  & 24.3 & 71.4  & 7.0   & 9.1  & 38,6  & 0.9   & 27.1 & 55.9  & 15.9  & 22.0 & 60.2  & 9.7   \\
                          & R50-MNv2  & 26.1 & 70.4  & 7.0   & 10.2 & 30.7  & 1.7   & 27.5 & 61.4  & 20.7  & 19.7 & 54.2  & 7.1   \\ \hline
\multirow{5}{*}{GraphKD}  & R50-R18   & 28.0 & 72.8  & 6.7   & 26.5 & 52.1  & 22.3  & 33.4 & 67.2  & 22.7  & 42.1 & 80.5  & 36.3  \\
                          & R101-R50  & 88.6 & 97.0  & 94.5  & 35.0 & 51.0  & 39.0  & 78.3 & 86.0  & 84.7  & 65.0 & 86.7  & 74.3  \\
 &
  R152-R101 &
  \textbf{88.8} &
  \textbf{97.0} &
  \textbf{94.7} &
  \textbf{41.9} &
  \textbf{58.9} &
  \textbf{42.0} &
  \textbf{79.7} &
  \textbf{87.4} &
  \textbf{86.0} &
  \textbf{68.9} &
  \textbf{89.0} &
  \textbf{78.0} \\
                          & R101-EB0  & 27.6 & 72.2  & 7.1   & 12.6 & 40.7  & 1.6   & 33.1 & 65.9  & 25.9  & 28.9 & 68.1  & 13.9  \\
                          & R50-MNv2  & 28.2 & 72.8  & 7.2   & 14.9 & 40.2  & 2.5   & 37.5 & 71.1  & 29.0  & 23.6 & 62.4  & 7.3   \\ \hline
\end{tabular}
}
\end{table}

It is worth noting that, the performance of the feature-based method (ReviewKD \cite{chen2021distilling}) is better compared to the logit-based method as the logits are like labels. That's why this NKD \cite{chi2023normkd} will be effective for the classification task not for the object detection task. On the other hand, when we combine the logits and features in SimKD \cite{chen2022knowledge} it outperforms the individuals as it gets the class-level information along with the features. Lastly, GraphKD creates nodes from the anchor boxes and also performs the node indexing which gets rid of the two complex task feature matching and logit matching through an easier node matching, which affects performance gain.

%% file: tex/conclusion.tex
In this paper, we have presented \emph{GraphKD}, a graph-based knowledge distillation strategy for document object detection to optimize the number of parameters so that, we use them in edge devices. According to our best knowledge, this is the first time, we have explored knowledge distillation for DOD where we used RoI pooled features for structured graph creation and performed an effective sample mining strategy to reduce the text bias. We also performed node-to-node distillation based on the similarity (using Cosine distance) and edge-to-edge distillation based on their covariance (i.e. Mahalanobis distance). From the results, it can be concluded that we are successful in reducing the number of parameters however, it also affects the performance ($\approx 7\%$ for PublayNet and DoclayNet and $\approx 13\%$ for PRIMA compared to SwinDocSegmenter). Also, we are unable to incorporate Transformers as a backbone in order to perform cross-architecture distillation (Transformers to ResNet) effectively due to their different data handling mechanism. It will be our future perspective of this work.

%% file: tex/supply.tex
\title{GraphKD: Exploring Knowledge Distillation Towards Document Object Detection with Structured Graph Creation}
\titlerunning{GraphKD}
\author{Supplementary}
\institute{Material}
\maketitle

\section{Implementation Details}
We train \emph{GraphKD} with an Adam optimizer with an initial learning rate of $1e-5$ with a cosine annealing scheduler of 5000 cycles utilizing weight decay of $1e-6$. To accumulate the final model we train for 90K iterations with a learning rate reduction by an order of magnitude 10 in the range of 70K to 80K. We prepared all models on NVIDIA A40 GPU of 48G RAM with 1 day of training utilizing stochastic learning. We utilize Pytorch and Detectron2 to build this framework. We need to tune three hyperparameters: namely edge distance for node merging, classification loss threshold during node indexing, and the covariance threshold during nude creation. We select those values as 0.1, 0.6, and 0.8 respectively. Also, in the graph distillation loss we need to tune $\lambda_1$, $\lambda_2$, and $\lambda_3$ in order to balance all the loss terms via regularization, we selected them as 0.001, 0.008, and, 0.003 respectively. (Note: all these hyperparameter values have been selected via hyperparameter grid search).

\section{A deep dive into the dataset description}
\begin{figure}[!htbp]
\centering
\subfloat[PubLayNet]{\includegraphics[width=0.25\textwidth]{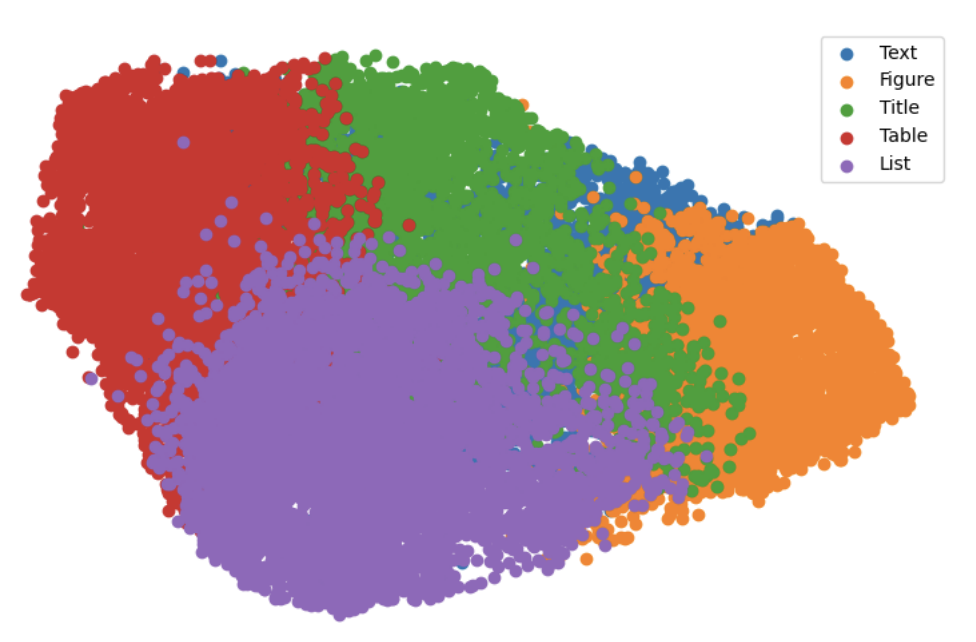}}\hfill
\subfloat[PRIMA]{\includegraphics[width=0.25\textwidth]{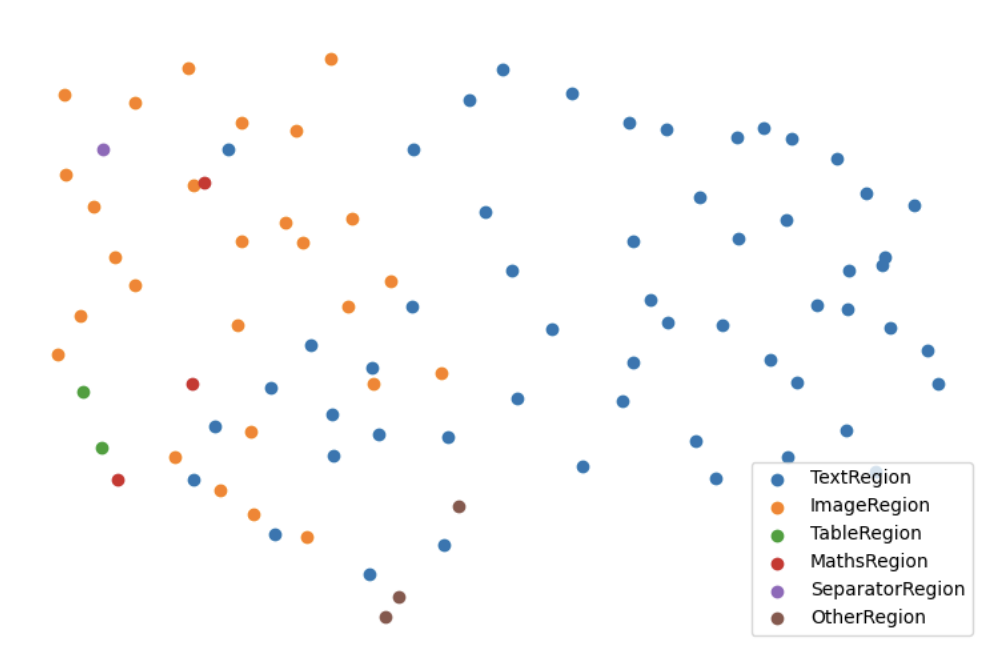}}\hfill
\subfloat[HJ]{\includegraphics[width=0.25\textwidth]{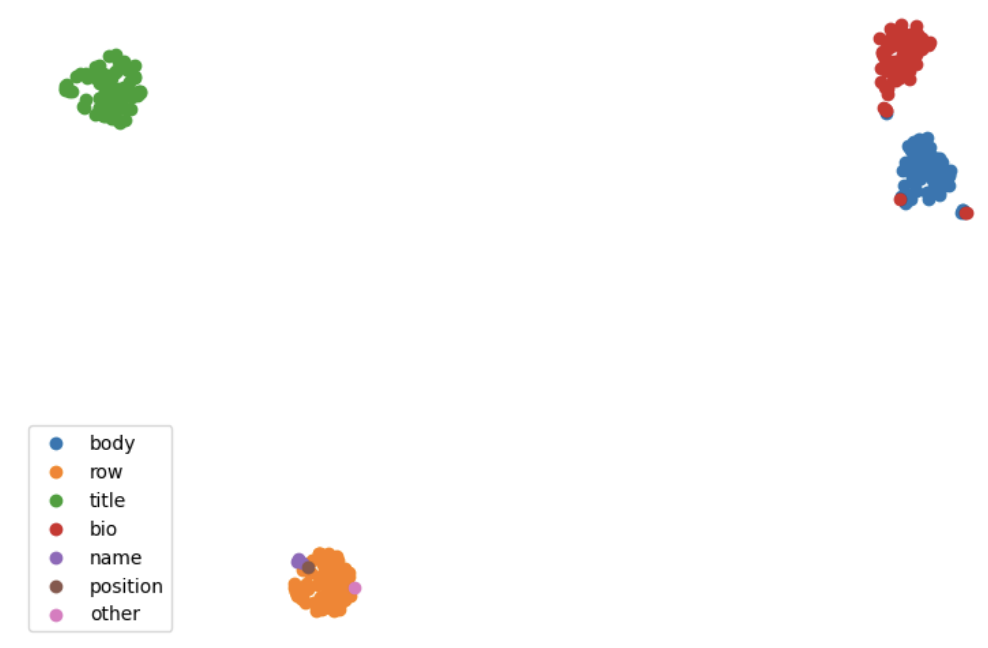}}\hfill
\subfloat[DocLayNet]{\includegraphics[width=0.25\textwidth]{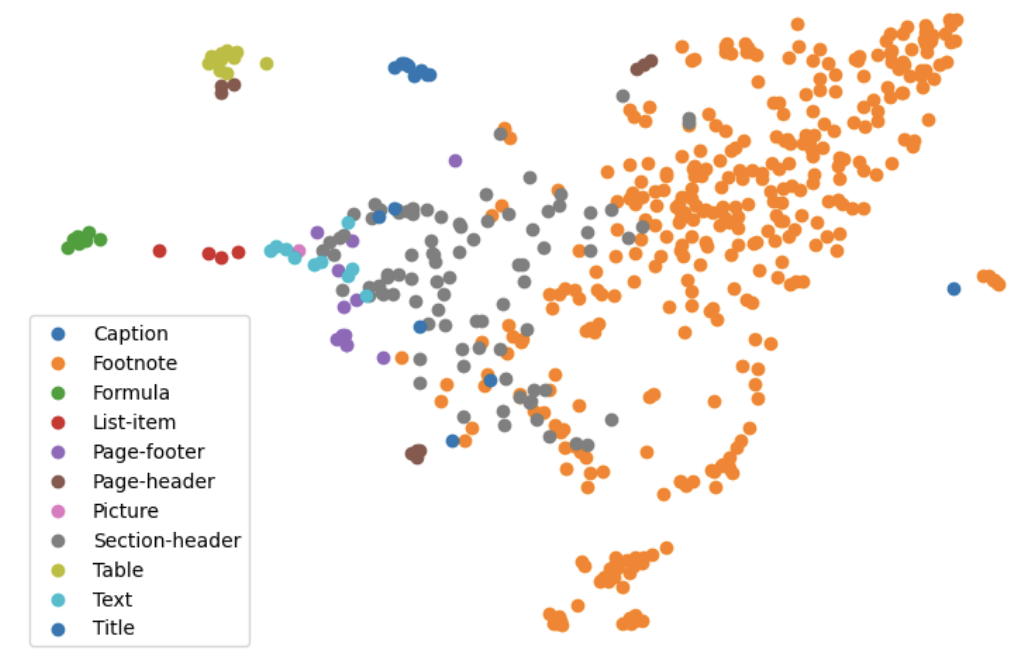}}
\caption{Understanding global relationship of the class instances through UMAP.}\label{fig:umap}
\end{figure}

From Table 1 of the main paper of \emph{GraphKD}, it has been observed that PubLayNet \cite{zhong2019publaynet} has the highest number of training/validation instances while PRIMA \cite{clausner2019icdar2019} has the lowest. On the other hand, DocLayNet \cite{pfitzmann2022doclaynet} has a large number of class instances although the data points are not well distributed. Besides that, it is important to understand the local and global relationships between the class instances. In order to identify that, we perform a UMAP visualization over the validation set as depicted in \cref{fig:umap}.

From the \cref{fig:umap} it has been observed that PubLayNet \cite{zhong2019publaynet} has a very dense data distribution among all the 5 classes (which helps to gain the instance segmentation performance) while PRIMA \cite{clausner2019icdar2019} suffers from the scarcity of the data problem. Other the other hand, in the historical Japanese dataset \cite{shen2020large}, classes are well separated, so it is easier to instantly segment them even with the low number of data points. Similarly, the DocLayNet has a large inter-class variability, with a large bias to the "list-item" and "text".

\section{Graph creation without node indexing}
\begin{figure}[h]
\centering
\subfloat[Graph Creation on DocLayNet]{\includegraphics[width=\textwidth]{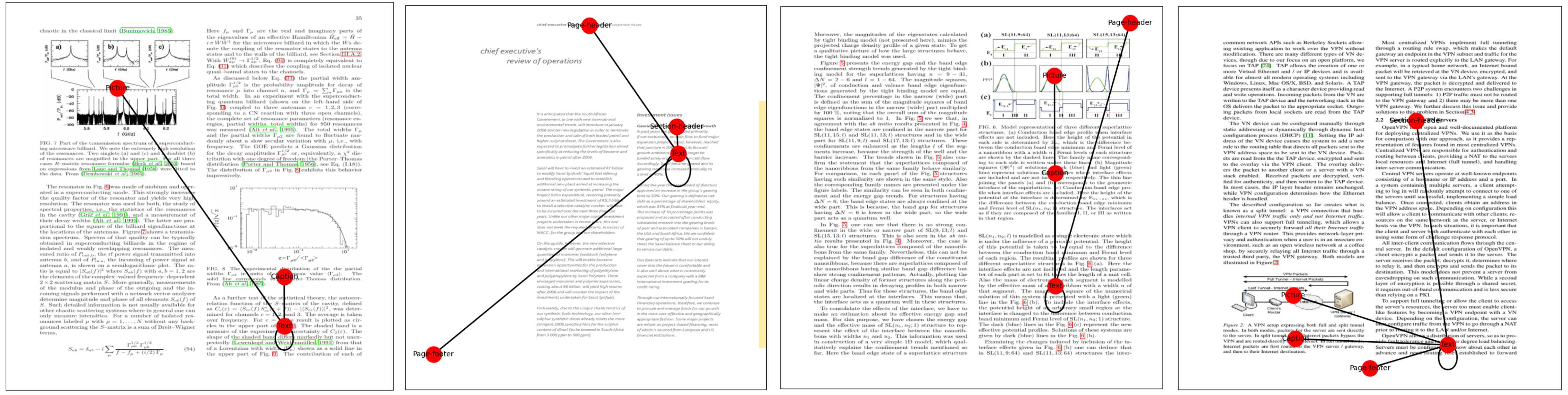}}\hfill \\
\subfloat[Graph Creation on PRIMA]{\includegraphics[width=\textwidth]{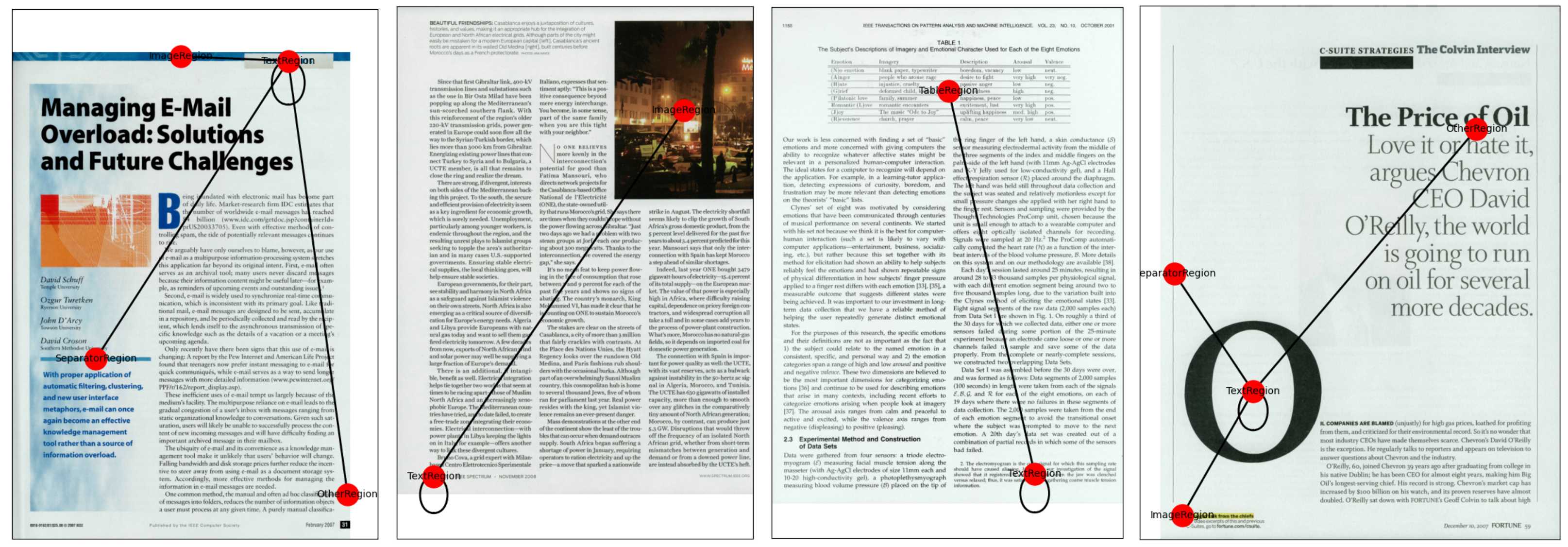}}\hfill
\caption{\textbf{Graph creation without node indexing:} Here one node represents the whole instances of each class as the nodes are developed on the feature embedding space.}\label{Fig:visgraph}
\end{figure}

We have discussed how we merge the text instances and put an index to each of the "text" nodes before merging in order to correctly localize the text region at the later stage in the object detection. In order to, perform node indexing we need to perform Object Sample Mining \cite{mooney2005mining}. We have identified the importance of this step by experimenting with the structured graph creation without node indexing.  \cref{Fig:visgraph} shows the effect of structured graph creation without node indexing.

It has been observed in \cref{Fig:visgraph}, that the obtained graph doesn't preserve the hierarchical structure of the documents which leads to poor object detection performance.

\section{Some qualitative insights}
\begin{figure}[!htbp]
\centering
\subfloat[R101-EB0]{\includegraphics[width=0.5\textwidth, height=3.0cm]{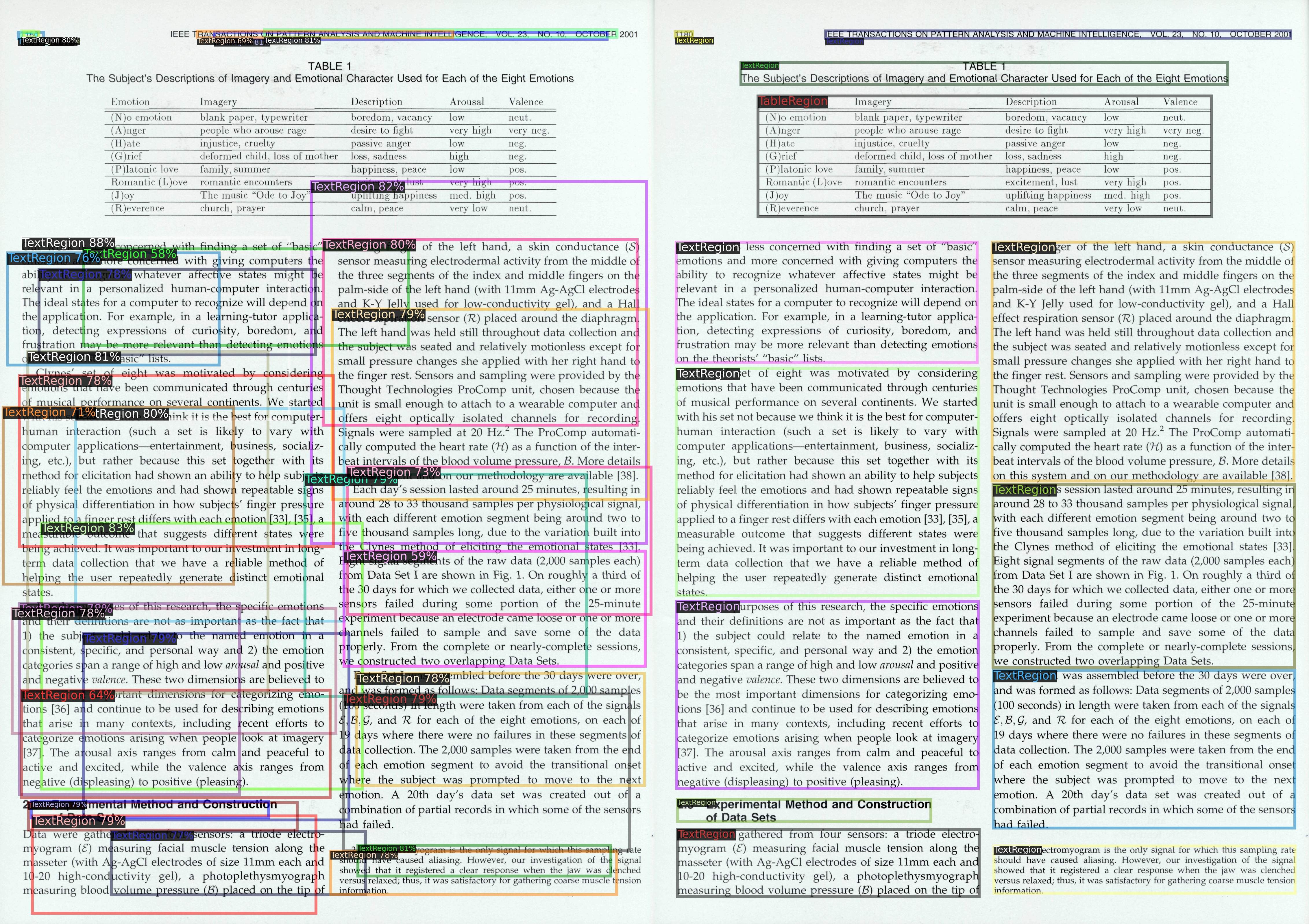}}\hfill
\subfloat[R101-EB0]{\includegraphics[width=0.5\textwidth, height=3.0cm]{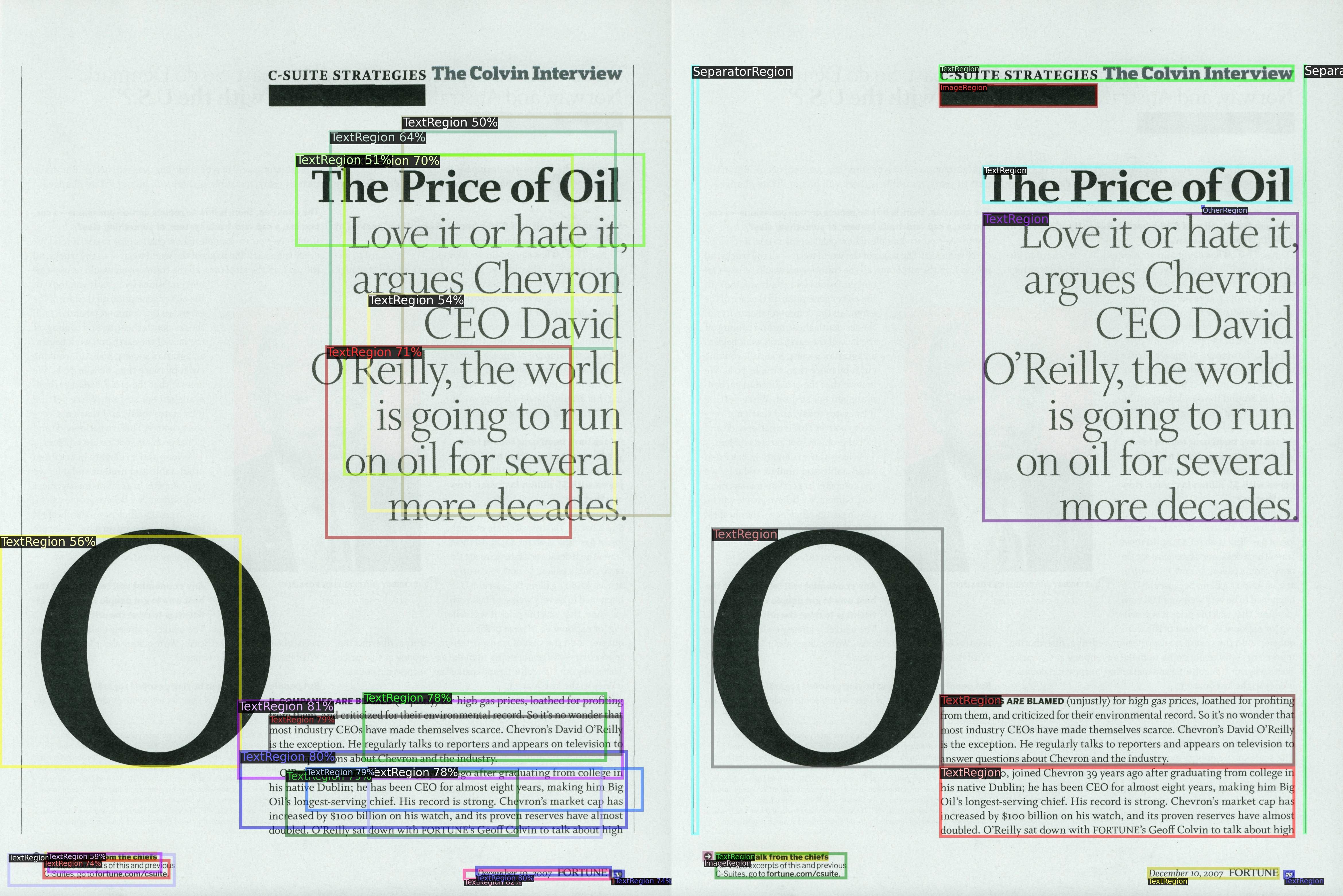}}\hfill \\
\subfloat[R50-MNv2]{\includegraphics[width=0.5\textwidth, height=3.0cm]{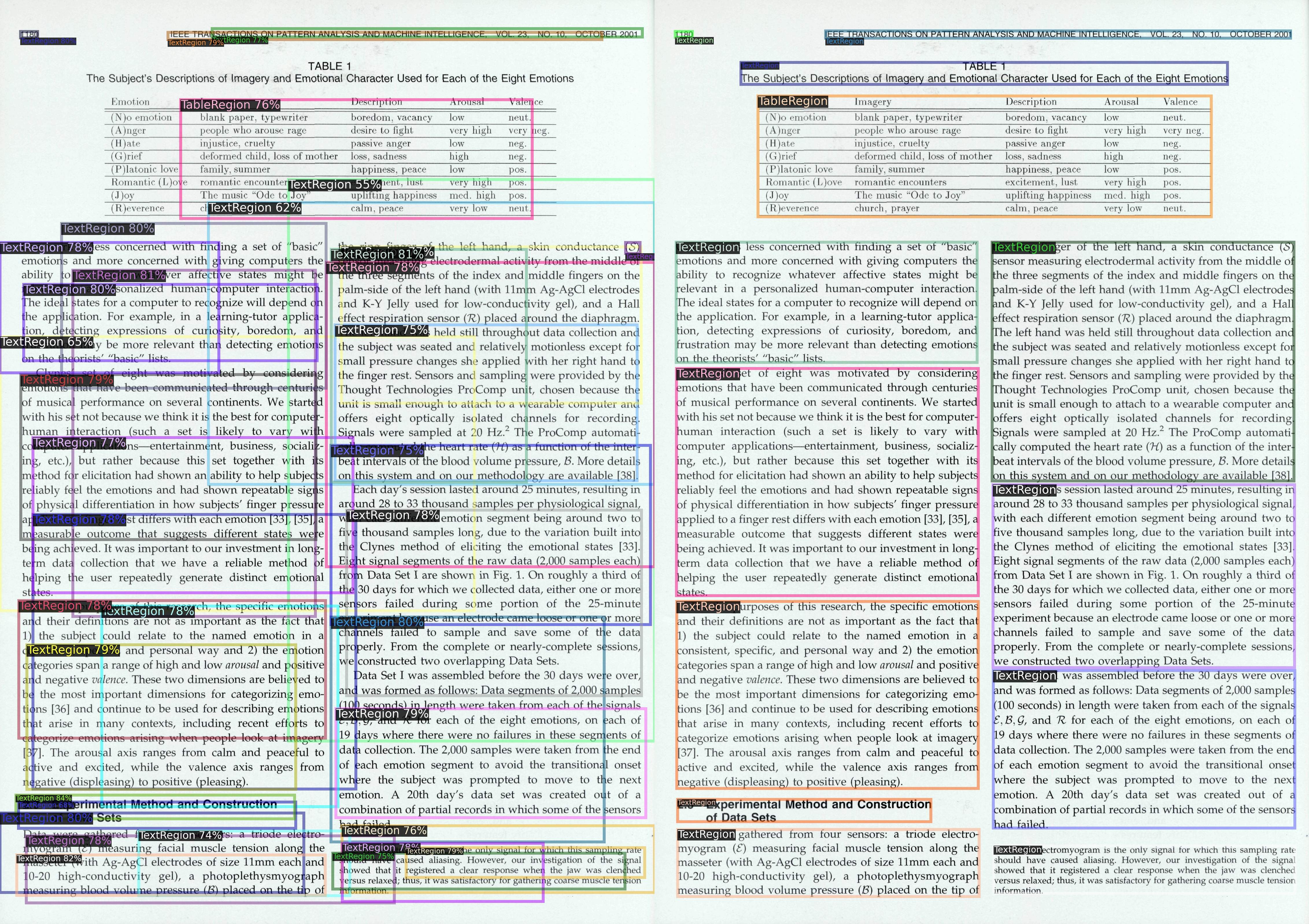}}\hfill
\subfloat[R50-MNv2]{\includegraphics[width=0.5\textwidth,height=3.0cm]{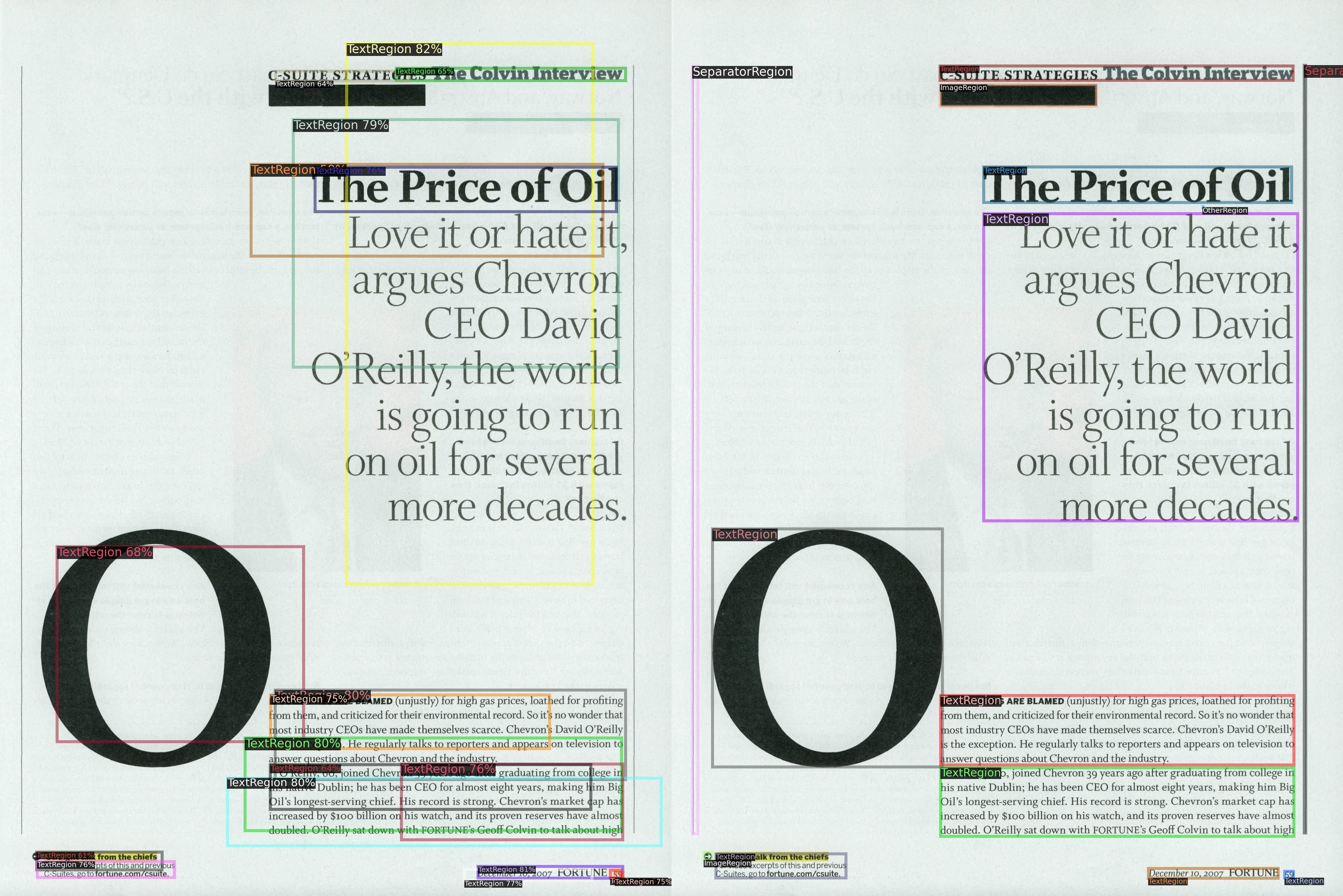}} \\
\subfloat[R50-R18]{\includegraphics[width=0.5\textwidth, height=3.0cm]{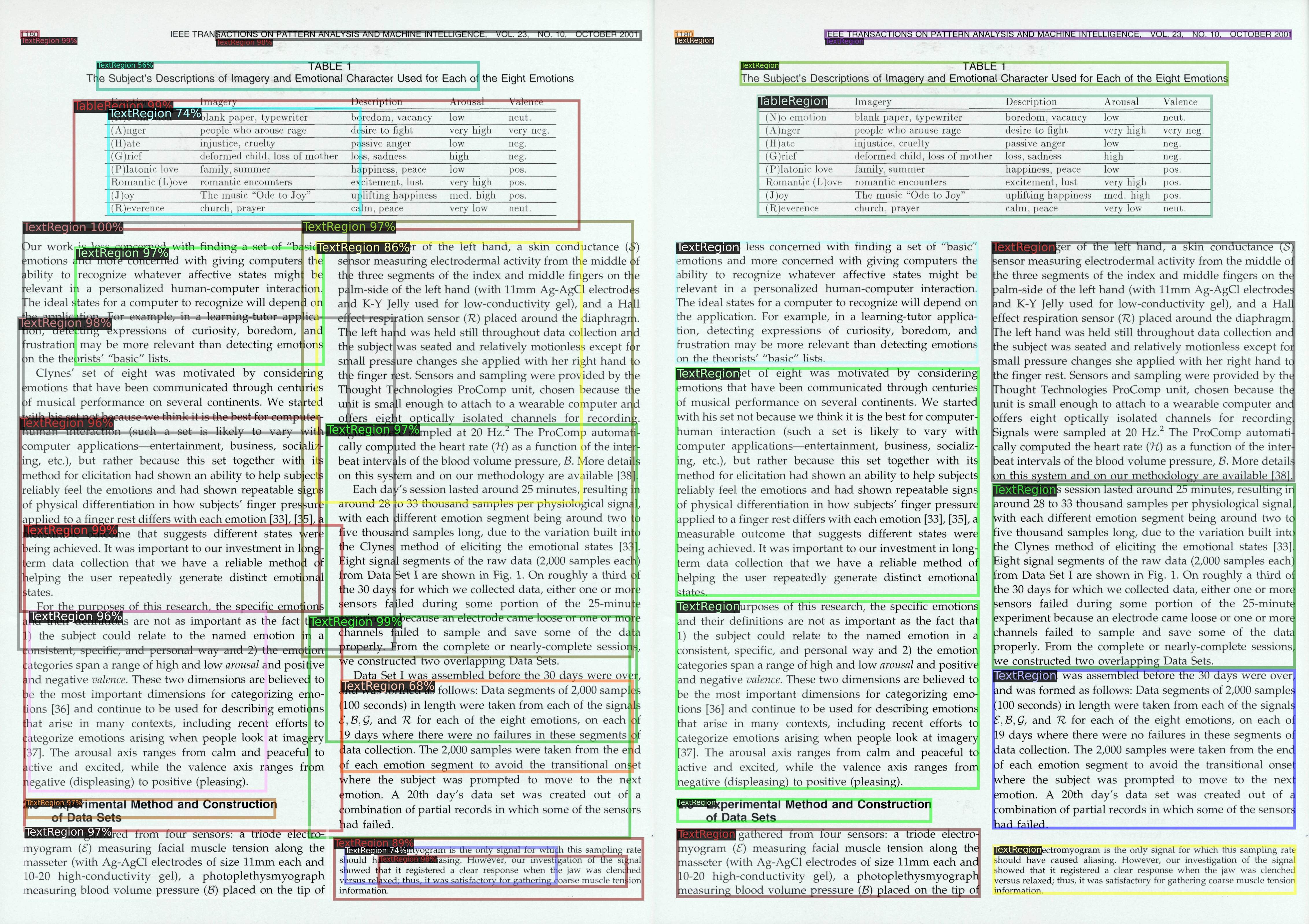}}\hfill
\subfloat[R50-R18]{\includegraphics[width=0.5\textwidth,height=3.0cm]{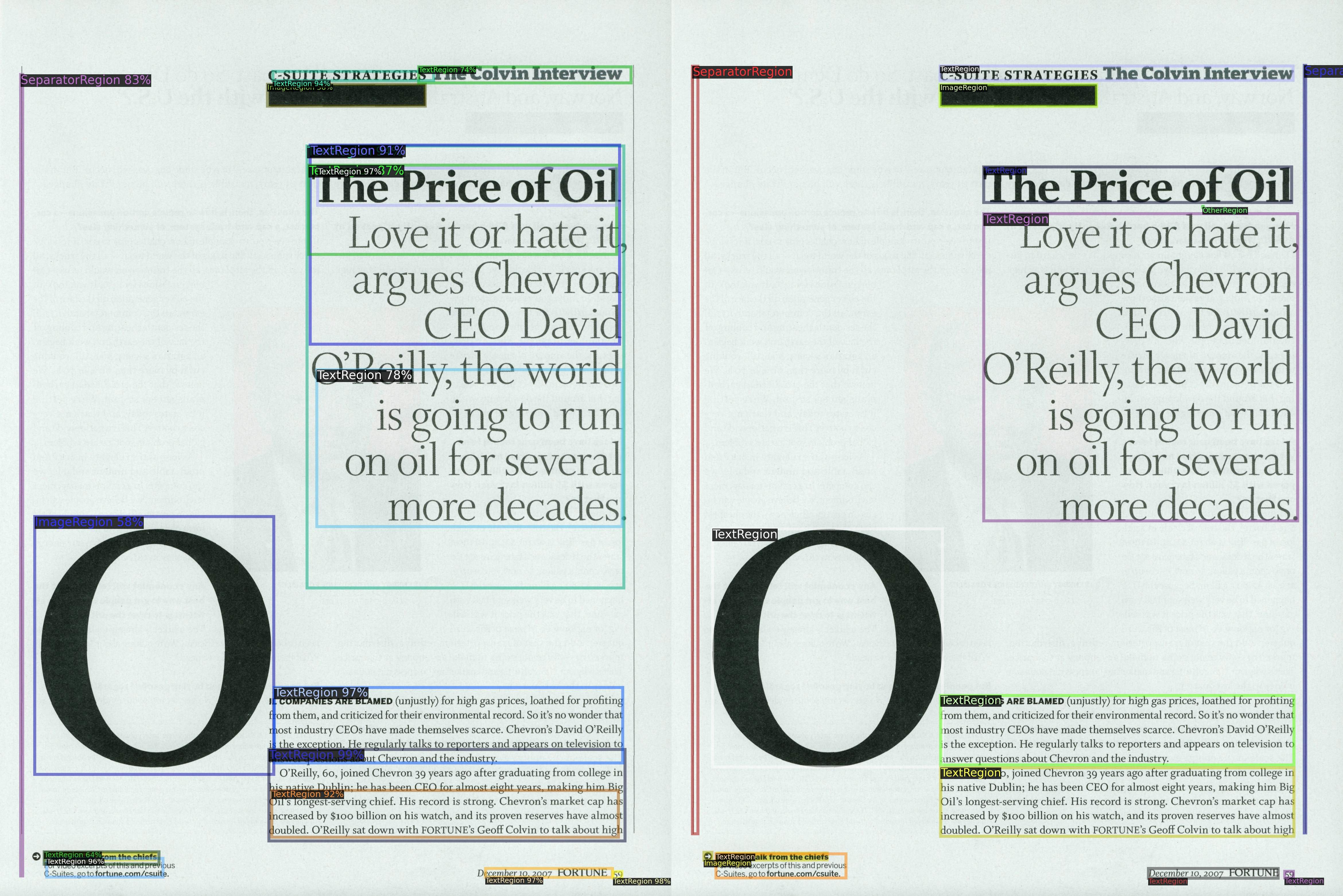}} \\
\subfloat[R101-R50]{\includegraphics[width=0.5\textwidth, height=3.0cm]{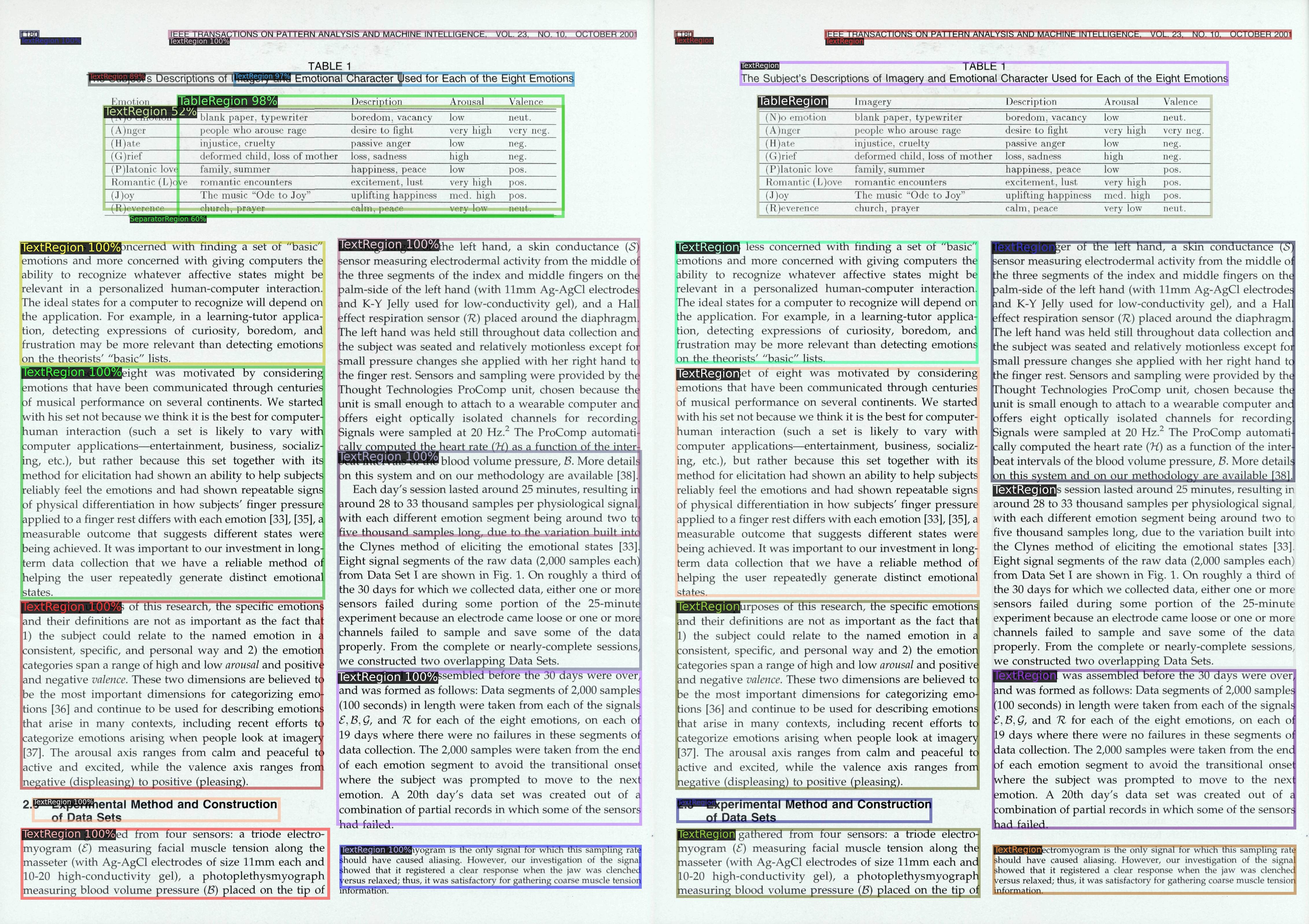}}\hfill
\subfloat[R101-R50]{\includegraphics[width=0.5\textwidth,height=3.0cm]{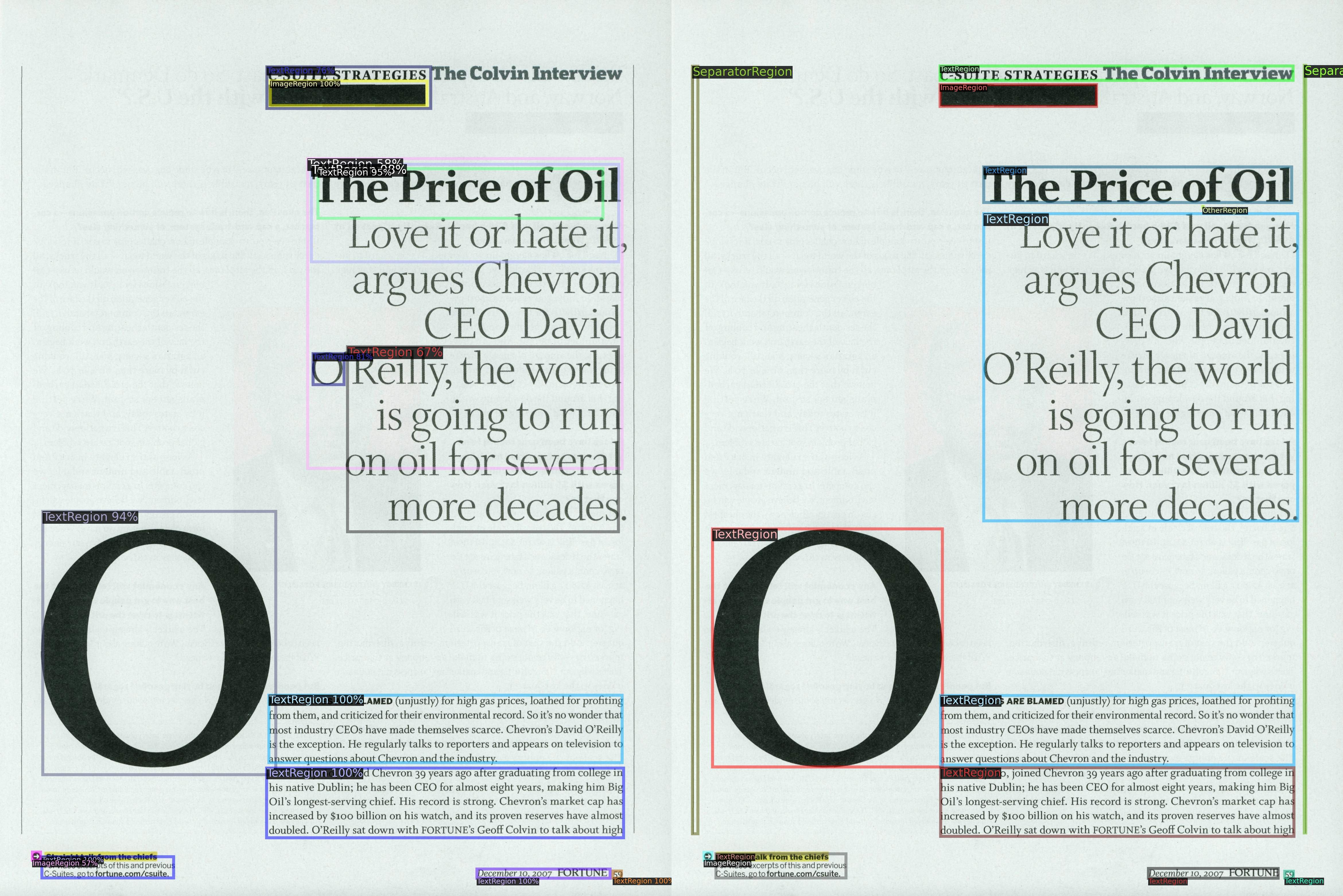}} \\
\subfloat[R152-R101]{\includegraphics[width=0.5\textwidth, height=3.0cm]{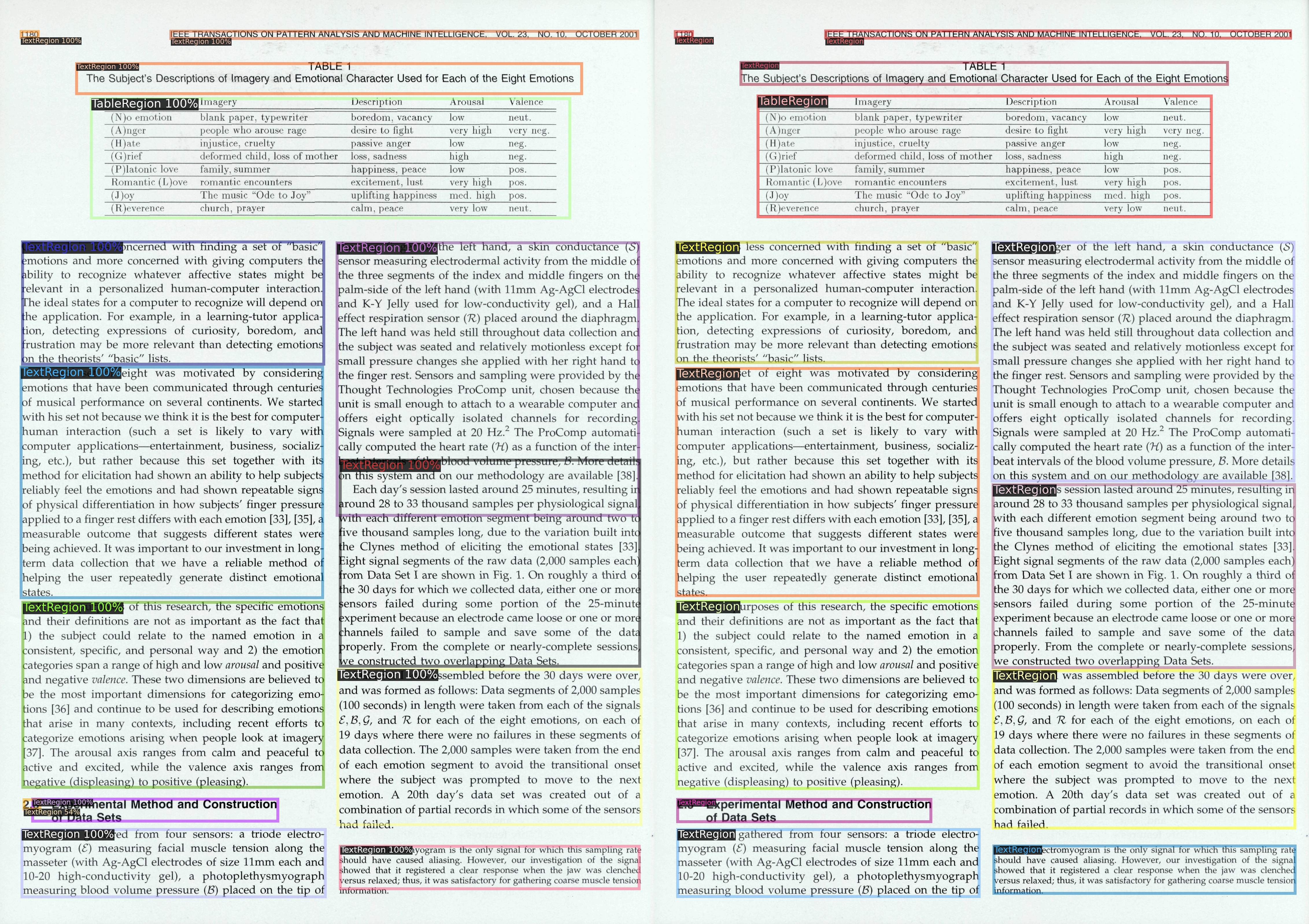}}\hfill
\subfloat[R152-R101]{\includegraphics[width=0.5\textwidth,,height=3.0cm]{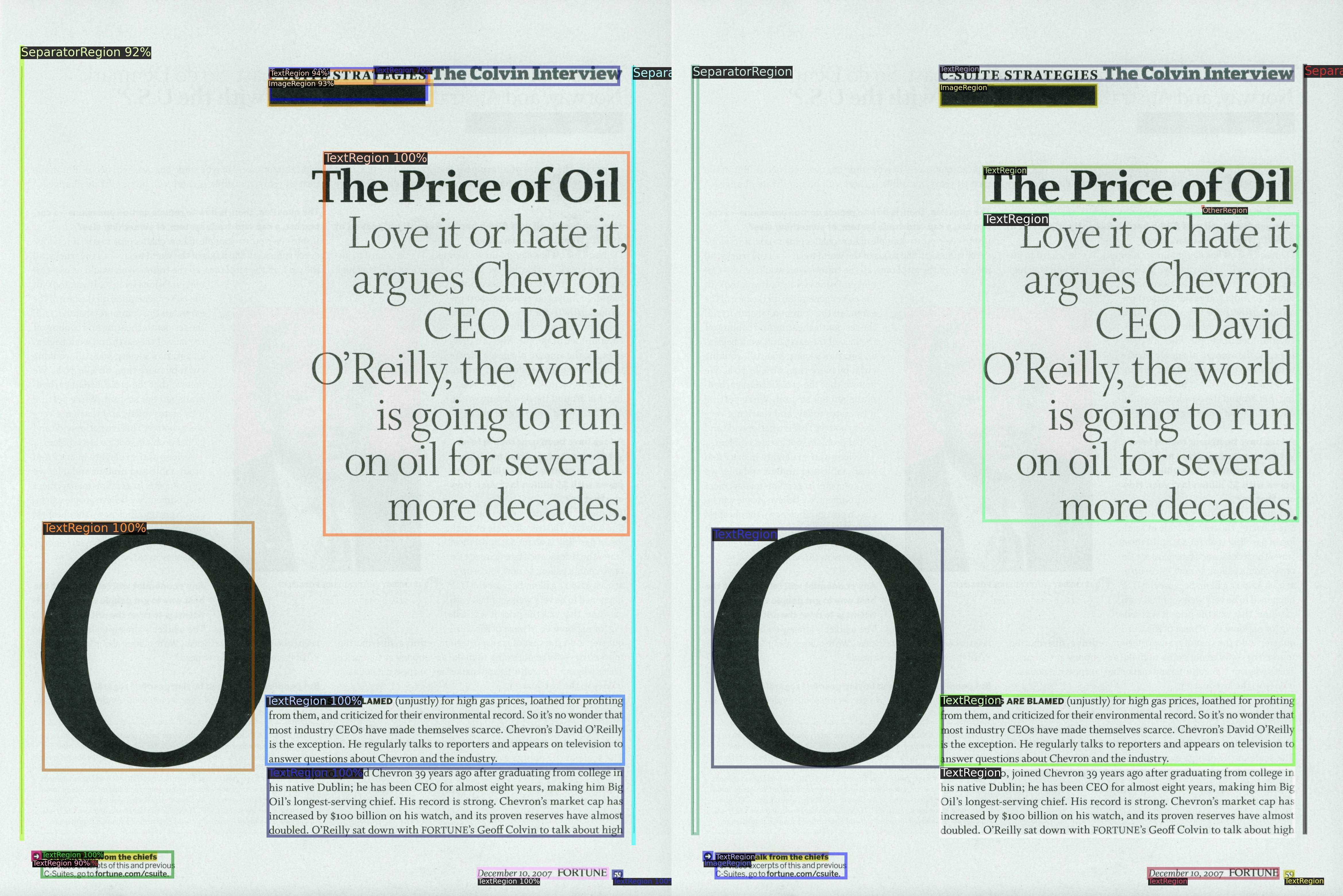}}
\caption{Qualitative analysis with various distilled networks on PRIMA dataset (left: predicted; right: ground-truth).}\label{fig:qual}
\end{figure}

Whenever we discuss knowledge distillation, we mainly focus on how to reduce no. of model parameters without affecting the object detection performance (i.e. mostly on mean average precision (mAP)). However, observing how the distilled networks perform on document images is always interesting. \cref{fig:qual} depicts some qualitative examples with all the distilled networks.

In the first example of \cref{fig:qual}(a) distilled efficient net B0 unable to detect the "Table". Similarly, in \cref{fig:qual}(b) it fails to detect the two "Separator" regions. On the other hand, distilled mobilenetv2 produces lots of false positives in \cref{fig:qual}(c) and is also not able to detect any of the "Separator" regions in \cref{fig:qual}(d). A significant with distilled ResNet18 has been observed which reduces a significant percentage of false positives in \cref{fig:qual}(e) and is also able to detect one of the "Separator" Region in \cref{fig:qual}(f). With ResNet50 there is only one false positive ("Table" region is also detected as "Text" region) in \cref{fig:qual}(g) however, none of the "Separator" region has been detected in \cref{fig:qual}(h). Only distilled ResNet101 is able to replicate the ground-truth in both the example which also validate our quantitative analysis.

\section{Performance of ResNets with Supervised Learning}
In Section 4.2 we have already observed the distillation performance of various convolution architectures and observed how distilled ResNet50, and ResNet101 provide comparable performance with large-scale transformer-based approaches. However, It is worth examining the performance of each feature extraction backbone for complex document object detection with supervised training. \cref{tab:s1} depicts the performance of supervised training of all the networks on four competitive benchmarks.

\begin{table}[!htbp]
\centering
\caption{Performance of various convolution networks with supervised training}
\label{tab:s1}
\resizebox{\textwidth}{!}{
\begin{tabular}{c|ccc|ccc|ccc|ccc}
\hline
\multirow{2}{*}{Backbone} &
  \multicolumn{3}{c|}{PublayNet \cite{zhong2019publaynet}} &
  \multicolumn{3}{c|}{PRIMA \cite{clausner2019icdar2019}} &
  \multicolumn{3}{c|}{HJ \cite{shen2020large}} &
  \multicolumn{3}{c}{DocLayNet \cite{pfitzmann2022doclaynet}} \\ \cline{2-13} 
     & AP   & AP@50 & AP@75 & AP   & AP@50 & AP@75 & AP   & AP@50 & AP@75 & AP   & AP@50 & AP@75 \\ \hline
R18  & 24.7 & 70.9  & 12.3  & 22.1 & 42.1  & 17.3  & 29.7 & 60.1  & 21.8  & 32.7 & 70.8  & 30.1  \\
R50  & 85.4 & 96.2  & 92.7  & 31.3 & 46.7  & 36.2  & 75.6 & 82.3  & 80.4  & 62.1 & 87.9  & 46.3  \\
R101 & 86.7 & 96.9  & 93.1  & 39.7 & 57.8  & 40.5  & 76.9 & 87.1  & 85.7  & 64.9 & 88.4  & 76.2  \\
R152 &
  \textbf{90.2} &
  \textbf{98.7} &
  \textbf{95.1} &
  \textbf{42.8} &
  \textbf{60.2} &
  \textbf{44.9} &
  \textbf{80.2} &
  \textbf{89.9} &
  \textbf{87.2} &
  \textbf{69.1} &
  \textbf{90.8} &
  \textbf{80.1} \\
EB0  & 23.7 & 69.2  & 17.1  & 11.1 & 30.8  & 1.2   & 30.1 & 62.9  & 23.9  & 26.9 & 58.1  & 11.7  \\
MNv2 & 24.2 & 70.8  & 21.1  & 12.6 & 35.9  & 2.0   & 34.8 & 70.5  & 26.1  & 20.6 & 56.4  & 6.3   \\ \hline
\end{tabular}
}
\end{table}

It has been observed that all the trained in a supervised manner obtained poor performance compared to the distilled one as the teacher model is typically a larger and more powerful model (e.g. ResNet152) that has learned to capture intricate patterns in the data. By distilling its knowledge into a smaller model, we effectively compress the information, making it easier to transfer and generalize to new data. 

\section{Failure case: cross architecture distillation}
The motivation for using graphs is to transfer knowledge more effectively by reducing the feature alignment, layer dimension check, and so on. Similarly, we also aim to perform cross-architecture distillation from transformers (ViT-B) to ResNet (R101). The performance has been reported in \cref{tab:s2}.

\begin{table}[!htbp]
\centering
\caption{Performance of knowledge distillation from ViT-Base to ResNet50}
\label{tab:s2}
\begin{tabular}{@{}cccccccc@{}}
\toprule
Dataset                    & backbone & AP   & AP@50 & AP@75 & APs  & APm  & APl  \\ \midrule
\multirow{2}{*}{PublayNet \cite{zhong2019publaynet}} & ViTB-R50 & 11.7 & 29.8  & 20.7  & 12.5 & 19.2 & 20.1 \\
                           & ViT-B    & 91.7 & 98.9  & 96.7  & 40.2 & 59.8 & 62.3 \\ \midrule
\multirow{2}{*}{PRIMA \cite{clausner2019icdar2019}}     & ViTB-R50 & 2.4  & 11.2  & 9.7   & 0.3  & 3.9  & 4.7  \\
                           & ViT-B    & 46.1 & 62.6  & 47.3  & 31.3 & 33.4 & 50.5 \\ \midrule
\multirow{2}{*}{HJ}        & ViTB-R50 & 9.8  & 22.8  & 16.7  & 8.2  & 15.1 & 16.7 \\
                           & ViT-B    & 81.2 & 89.7  & 84.1  & 36.8 & 55.7 & 56.8 \\ \midrule
\multirow{2}{*}{DoclayNet \cite{pfitzmann2022doclaynet}} & ViTB-R50 & 5.7  & 24.6  & 13.2  & 1.2  & 7.8  & 8.2  \\
                           & ViT-B    & 65.6 & 84.7  & 73.6  & 37.8 & 55.4 & 59.2 \\ \bottomrule
\end{tabular}
\end{table}
It has been observed that the performance gap between the teacher and student model is quite large due to feature misalignment between the transformer layer and the convolution layer. Also, the transformer used a self-attention mechanism and in Resnet we used Relu activation so the Region proposal is completely different for these two backbones (feature compression is not possible), so we cannot use shared RPN which leads to this feature misalignment and we cannot perform node indexing based on the teacher classification loss during distillation.